%% file: neurips_2026.tex
\documentclass{article}

    \PassOptionsToPackage{numbers, compress}{natbib}

 \usepackage[main, final]{neurips_2026}

\usepackage[utf8]{inputenc} 
\usepackage[T1]{fontenc}    
\usepackage{hyperref}       
\usepackage{url}            
\usepackage{booktabs}       
\usepackage{amsfonts}       
\usepackage{nicefrac}       
\usepackage{microtype}      
\usepackage[table]{xcolor}  

\usepackage{amsmath}
\usepackage{amsthm}

\usepackage{graphicx}
\usepackage{multirow}
\usepackage{makecell}

\usepackage{listings}
\title{Capturing In-Context Learning Dynamics with\\Task Operators}

\author{%
  Guangzhi Xiong \\
  University of Virginia \\
  \texttt{guangzhi@virginia.edu} \\
\And
  Zhenghao He \\
  University of Virginia \\
  \texttt{zhenghao@virginia.edu} \\
\And
  Bohan Liu \\
  University of Virginia \\
  \texttt{bohan@virginia.edu} \\
\And
  Sanchit Sinha \\
  University of Virginia \\
  \texttt{sanchit@virginia.edu} \\
\And
  Wenqian Ye \\
  University of Virginia \\
  \texttt{wenqian@virginia.edu} \\
\And
  Aidong Zhang \\
  University of Virginia \\
  \texttt{aidong@virginia.edu} \\
}

\begin{document}

\maketitle

\begin{abstract}
  \input{contents/abstract}
\end{abstract}

\input{contents/introduction}
\input{contents/related_work}
\input{contents/methodology}
\input{contents/experiments}
\input{contents/conclusion}

\begin{ack}
This work is supported in part by the US National Science Foundation (NSF) and the National Institute of Health (NIH) under grants 2616033, 2538206, 2529378, 2217071, 2213700, R01LM014012-01A1, and 1R01HL184168-01. Any recommendations expressed in this material are those of the authors and do not necessarily reflect the views of NIH or NSF.
\end{ack}

\bibliographystyle{plainnat}
\bibliography{references}


\clearpage
\appendix

\input{contents/appendix}



\end{document}

%% file: contents/abstract.tex
In-context learning (ICL) enables language models to perform new tasks from demonstrations without weight updates. However, every ICL inference requires processing the full set of examples, resulting in inefficient deployments, and how ICL works mechanistically is not fully understood. Prior work compresses ICL into fixed activation vectors extracted from specific layers or positions, but these input-independent interventions fail on complex tasks where the output depends on fine-grained interactions with the input. By analyzing the ICL forward pass, we show that each attention head's output is an affine transformation of its context-masked counterpart, and that the parameters of this transformation are empirically stable across samples for a given task. Building on this, we introduce Task Operator (TO), which replays this transformation as an analytically derived update to the attention output projection. Across lexical, algorithmic, and reasoning tasks, TO achieves the best overall performance among prior methods and substantially narrows the gap between zero-shot inference and ICL. We further show that the extracted knowledge concentrates in a task-specific sparse circuit across layers and positions, and that averaging operators from disjoint demonstration batches enables effective many-shot scaling without expanding the context window.
Our code is available at \url{https://github.com/gzxiong/task_operator}.

%% file: contents/introduction.tex
\section{Introduction}

In-context learning (ICL) enables large language models (LLMs) to quickly adapt to new tasks by providing input-output demonstrations in the context \citep{brown2020language}. This allows LLMs to improve performance on tasks where they fail in the zero-shot learning (ZSL) setting, without any additional training. However, ICL requires processing all demonstrations for the inference of each new input, which can be computationally expensive and inefficient even with KV caching \cite{pope2023efficiently}.
Therefore, it is important to understand the mechanisms underlying ICL and to develop methods that can distill the ICL representation into a compact form, enabling efficient replay at inference time without repeatedly processing demonstrations.

Recent approaches tackle this inefficiency by developing methods that compress ICL into lightweight, fixed representations, allowing models to bypass demonstrations and efficiently replay task knowledge during inference \citep{hendel2023context, todd2024function, liu2024incontext, postmus2024steering, li2024incontext}. These methods typically extract task-specific activation vectors from particular layers or token positions during ICL and inject them as fixed interventions during zero-shot inference. While effective for simple tasks with clear input-output mappings, these approaches fail when output generation requires complex multi-step reasoning over the input content, indicating that the activation vectors do not faithfully capture the knowledge encoded in the ICL process.

By analyzing how demonstrations reshape the model's internal computation, we show that at every attention head, the output under ICL is precisely an affine transformation of its context-masked counterpart, with the parameters of these transformations showing high consistency across samples (Appendix \ref{sec:appendix_stability}). With extracted parameters for all attention heads, we show that the effects of demonstrations can be exactly represented as an input-dependent reparameterization of the attention output projection.

Building on this analysis, we introduce the Task Operator (TO), a training-free method that captures ICL knowledge as affine transformation parameters and replays it during inference to dynamically update the attention outputs. Through experiments on four LLMs across eight tasks, we demonstrate that TO faithfully captures ICL knowledge and outperforms prior knowledge extraction methods overall when replaying the knowledge in zero-shot inference. Causal circuit analysis and ablation studies reveal that the key sites for storing ICL knowledge are sparse and task-specific, explaining why previous methods that extract knowledge from fixed locations fall short. Scaling analysis further demonstrates the effectiveness of TO under many-shot ICL, where knowledge is averaged from disjoint mini-batches of demonstrations, highlighting its scalability within a fixed context budget.
Our main contributions are:
\begin{itemize}
    \item We propose Task Operator (TO), a training-free method that captures ICL knowledge as an affine transformation of the attention output, enabling efficient and dynamic replay of ICL knowledge during zero-shot inference.
    \item Evaluation across four LLMs and eight tasks demonstrates that TO achieves the best overall performance, highlighting the importance of dynamic interventions for faithfully capturing and replaying ICL knowledge.
    \item Our analysis reveals the diverse and sparse circuits leveraged by ICL across different tasks, and shows that TO enables efficient many-shot scaling by averaging knowledge extracted from disjoint demonstration batches.
\end{itemize}

%% file: contents/related_work.tex
\section{Related Work}
\label{sec_related_work}

\paragraph{In-context learning mechanisms.}
In-context learning enables language models to perform new tasks from input-output demonstrations without parameter updates \citep{brown2020language}. 
Several theoretical accounts have been proposed to explain this behavior. 
\citet{xie2022explanation} interpret ICL as implicit Bayesian inference over latent document-level concepts induced by long-range coherence in pretraining data. 
\citet{dai2023gpt} connect transformer attention to gradient descent and argue that ICL implements implicit fine-tuning through meta-gradients computed from demonstrations. 
Mechanistic studies identify internal structures that support ICL. 
\citet{olsson2022context} link induction heads to the emergence of ICL during training, while \citet{yin2025which} show that few-shot ICL is driven primarily by function vector heads rather than induction heads, with relevant heads distributed across layers in a task-dependent manner. 
Recent work further studies how task representations emerge and what forms they can take. 
\citet{han2025emergence} propose an encoder-decoder account in which early layers encode task information and later layers decode it, while \citet{dong2026understanding} analyze the linear combination structure of task vectors and prove that additive task vectors are restricted to rank-one meta predictors. 
Our work builds on this line of inquiry by deriving an exact per-head affine form for the effect of demonstrations and using it to replay ICL behavior without reprocessing the demonstration context.

\paragraph{Compressing ICL into task representations.}
A parallel line of work seeks to compress ICL into reusable representations that can be applied without repeatedly processing demonstrations. 
\citet{ilharco2023editing} introduce task vectors in weight space, computed from the difference between fine-tuned and pre-trained model weights. 
\citet{hendel2023context} show that ICL creates analogous task vectors in activation space, where the hidden state at a separator token after demonstrations encodes a reusable task representation. 
\citet{todd2024function} extend this idea to function vectors by aggregating outputs from causally important attention heads, enabling transfer to zero-shot and natural text settings. 
Other training-free methods construct task representations from broader activation statistics. 
\citet{liu2024incontext} propose in-context vectors that use principal directions from demonstration hidden states. 
\citet{li2024incontext} introduce state vectors that concatenate intermediate activations and use momentum based optimization at test time, while \citet{li2025implicit} compress demonstrations into context vectors through mean shift in the residual stream. 
These methods provide efficient alternatives to full ICL, but they primarily replay task knowledge through fixed additive interventions. 
In contrast, task operators include both multiplicative scaling and additive context contributions at each attention head, allowing the replayed intervention to depend on the test-time hidden state.

\paragraph{Richer task vector and activation interventions.}
Recent methods address the limitations of fixed additive vectors by introducing input dependence, learned components, or richer intervention families. 
DyVec \citep{cai2025beyond} dynamically segments multi-head attention output projections for each input, but requires REINFORCE-based training. 
Adaptive Task Vectors \citep{kang2025adaptive} train a small language model to generate a task vector conditioned on each input, and In-Context Routing \citep{li2026train} trains an input-conditioned router that adjusts attention logits.
LIVE \citep{peng2024live} and LTV \citep{saglam2025learning} learn per-layer vectors via gradient optimization. 
Related work on activation steering also shows that additive interventions alone can be insufficient. 
\citet{singh2024representation} derive affine steering functions for residual stream interventions in concept erasure and bias mitigation. 
\citet{marshall2025refusal} show that refusal behavior is better characterized as an affine function of internal representations, while \citet{stoehr2024activation} propose activation scaling via element-wise multiplicative factors.
\citet{hernandez2024linearity} show that, for a subset of relations, relation decoding in language models is well approximated by an affine map derived from a first-order approximation of the model. In contrast, our per-head affine form is an exact identity for each input.
Conceptor-based methods use soft projection matrices for activation engineering \citep{postmus2024steering}, with subsequent work deriving compositional affine steering mechanisms \citep{abreu2024from}. 
These approaches show that richer interventions can improve model steering, but they either introduce additional learned or optimized components, or estimate steering transformations empirically from activation statistics rather than deriving an analytical knowledge extraction rule.
In contrast, our method derives input-dependent affine parameters directly from the ICL forward pass and replays them through dynamic updates to the attention output projection.

\paragraph{Mechanistic interpretability and circuit localization.}
Mechanistic interpretability aims to explain model behaviors in terms of internal components and their interactions. 
\citet{elhage2021mathematical} and \citet{olsson2022context} develop the conceptual framework of circuits as computational subgraphs within transformers. 
\citet{wang2023interpretability} present a detailed circuit analysis of indirect object identification in GPT-2, identifying attention heads organized into functional classes. 
\citet{geva2023dissecting} trace information flow for factual recall, showing how MLP layers enrich subject representations before attention heads extract relevant attributes. 
Causal intervention techniques, including activation patching \citep{vig2020investigating, meng2022locating} and path patching \citep{goldowskydill2023localizing}, provide tools for isolating the contributions of specific model components. 
Our work follows this tradition by using causal analysis to localize where ICL knowledge is stored, showing that the relevant sites form sparse and task-specific circuits, and using these localized effects to construct task operators that can be replayed during zero-shot inference.

%% file: contents/methodology.tex
\section{Task Operators in In-Context Learning}
\label{sec:method}

\begin{figure}[h!]
    \centering
    \includegraphics[width=\linewidth]{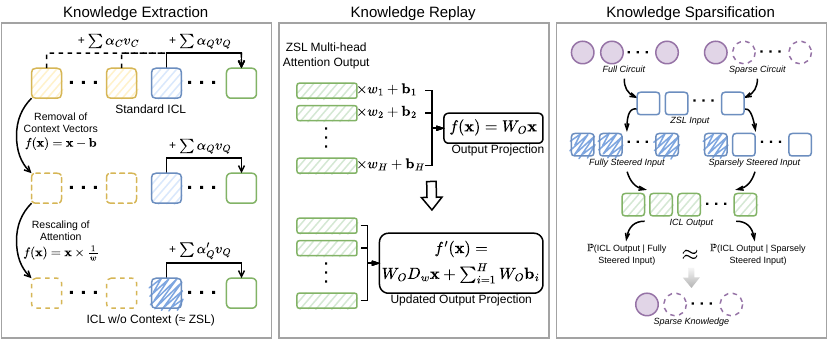}
    \caption{Overview of the Task Operator framework: extracting per-head affine parameters $(w, \mathbf{b})$ from ICL forward passes (\textit{left}), replaying them via an updated output projection at zero-shot inference (\textit{middle}), and sparsifying to the highest-importance sites (\textit{right}). The yellow blocks denote the context tokens, and the blue blocks denote the tokens of the input query.}
    \label{fig:framework}
\end{figure}

We introduce Task Operator (TO), a method for transferring in-context learning (ICL) behavior to zero-shot inference via dynamic interventions at multiple sites throughout the model.
Our approach consists of three main components:
First, we derive the affine structure underlying ICL's effect on attention heads and show how to extract the relevant parameters from ICL forward passes (Section~\ref{sec:extraction}).
Second, we describe how the extracted operator is replayed during zero-shot inference (Section~\ref{sec:inference}).
Third, we present a sparsification procedure that identifies and retains only the most important (layer, token) sites for intervention (Section~\ref{sec:sparsification}).
Figure~\ref{fig:framework} illustrates the overall framework.

\subsection{Knowledge Extraction}
\label{sec:extraction}

At layer $\ell$, for a query token at position $t$, each attention head $h$ computes attention probabilities $\alpha_{t,i}^{(\ell,h)} = \exp(s_{t,i}^{(\ell,h)}) / Z_t^{(\ell,h)}$ over all preceding positions $i \leq t$, where $s_{t,i}^{(\ell,h)}$ are the pre-softmax scores and $Z_t^{(\ell,h)} = \sum_{i \leq t} \exp(s_{t,i}^{(\ell,h)})$.
The per-head output $\mathbf{u}_t^{(\ell,h)} = \sum_{i \leq t} \alpha_{t,i}^{(\ell,h)} \, \mathbf{v}_i^{(\ell,h)}$ aggregates value vectors weighted by these probabilities.
The heads are concatenated and projected through the output projection, $\mathrm{Attn}_t^{(\ell)} = \mathbf{W}_O^{(\ell)} \mathbf{U}_t^{(\ell)} + \mathbf{b}_O^{(\ell)}$, where $\mathbf{U}_t^{(\ell)} = [\mathbf{u}_t^{(\ell,1)}; \ldots; \mathbf{u}_t^{(\ell,H)}] \in \mathbb{R}^d$ and each head has dimension $d_h = d / H$.

\paragraph{Affine decomposition.}
For clarity, we fix a layer $\ell$, a head $h$, and a query position $t$, and omit these indices in Equations~\eqref{eq:head-decomp}--\eqref{eq:per-head-identity}.
Let $\mathcal{C}$ denote the context token positions (the demonstration examples) and $\overline{\mathcal{C}} = \{i : i \leq t, \, i \notin \mathcal{C}\}$ the non-context positions accessible from the query position.
The head output decomposes as
\begin{equation}
\label{eq:head-decomp}
\mathbf{u}
\;=\; \underbrace{\textstyle\sum_{i \in \overline{\mathcal{C}}} \alpha_{i} \, \mathbf{v}_i}_{\text{non-context contribution}}
\;+\; \underbrace{\textstyle\sum_{i \in \mathcal{C}} \alpha_{i} \, \mathbf{v}_i}_{\text{context contribution} \;=:\; \mathbf{b}}.
\end{equation}
Define the context-masked head output, obtained by renormalizing over non-context positions only, as
$\mathbf{u}' = \sum_{i \in \overline{\mathcal{C}}} \alpha'_{i} \, \mathbf{v}_i$,
where $\alpha'_{i} = \exp(s_{i}) / Z'$ and $Z' = \sum_{i \in \overline{\mathcal{C}}} \exp(s_{i})$.
The non-context contribution in \eqref{eq:head-decomp} can then be rewritten as
\begin{equation}
\label{eq:rescale}
\sum_{i \in \overline{\mathcal{C}}} \alpha_{i} \, \mathbf{v}_i
= \frac{Z'}{Z} \sum_{i \in \overline{\mathcal{C}}} \alpha'_{i} \, \mathbf{v}_i
= w \, \mathbf{u}',
\end{equation}
where $w \equiv Z' / Z \in (0, 1]$ is the fraction of softmax mass on non-context positions.
Combining Equations~\eqref{eq:head-decomp} and~\eqref{eq:rescale} gives the per-head affine identity
\begin{equation}
\label{eq:per-head-identity}
\mathbf{u} = w \, \mathbf{u}' + \mathbf{b}.
\end{equation}
In words, the head output under ICL equals its context-masked output, rescaled by the attention mass that remains on non-demonstration tokens, plus the value contribution retrieved from the demonstrations.
This identity is exact for each input and holds for arbitrary attention patterns and context spans.

\paragraph{Output projection reparameterization.}
Restoring the indices $(\ell, h, t)$, let $\mathbf{D}_t^{(\ell)} = \mathrm{diag}\!\big(\underbrace{w_t^{(\ell,1)}, \ldots, w_t^{(\ell,1)}}_{d_h}, \ldots, \underbrace{w_t^{(\ell,H)}, \ldots, w_t^{(\ell,H)}}_{d_h}\big) \in \mathbb{R}^{d \times d}$
and $\mathbf{B}_t^{(\ell)} = [\mathbf{b}_t^{(\ell,1)}; \ldots; \mathbf{b}_t^{(\ell,H)}] \in \mathbb{R}^d$.
Concatenating Equation~\eqref{eq:per-head-identity} across heads and applying $\mathbf{W}_O^{(\ell)}$ gives
\begin{equation}
\label{eq:reparameterization}
\mathrm{Attn}_t^{(\ell)}
= \mathbf{W}_O^{(\ell)} \big( \mathbf{D}_t^{(\ell)} \, \mathbf{U}_t^{\prime(\ell)} + \mathbf{B}_t^{(\ell)} \big)  + \mathbf{b}_O^{(\ell)}
= \underbrace{\mathbf{W}_O^{(\ell)} \mathbf{D}_t^{(\ell)}}_{\widetilde{\mathbf{W}}_{O,t}^{(\ell)}} \, \mathbf{U}_t^{\prime(\ell)} \;+\; \underbrace{\mathbf{W}_O^{(\ell)} \mathbf{B}_t^{(\ell)} + \mathbf{b}_O^{(\ell)}}_{\widetilde{\mathbf{b}}_{O,t}^{(\ell)}}.
\end{equation}
The modified weight $\widetilde{\mathbf{W}}_{O,t}^{(\ell)} = \mathbf{W}_O^{(\ell)} \mathbf{D}_t^{(\ell)}$ is a diagonal scaling of $\mathbf{W}_O$ on its input side, where each of the $H$ head channels can be independently scaled by its corresponding $w_t^{(\ell,h)}$.
Since $\mathbf{D}_t^{(\ell)}$ and $\mathbf{B}_t^{(\ell)}$ are computed from each input, Equation~\eqref{eq:reparameterization} expresses the ICL effect as an input-dependent reparameterization of the output projection.

\paragraph{Extraction procedure.}
Since the distribution of ICL-critical positions may vary across tasks \citep{sia2024where}, we extract from all token positions across all layers rather than selecting a fixed subset.
When the ICL effect is negligible at a site, the parameters are close to identity ($w \approx 1$, $\mathbf{b} \approx \mathbf{0}$), thus having minimal impact when replayed at test time.
Given an ICL prompt, we run the LLM and record pre-softmax attention scores at each layer.
For each site $(\ell, t)$, we compute the per-head scaling factor via the numerically stable form
\begin{equation}
\label{eq:w-stable}
w_t^{(\ell,h)} = 1 - \exp\!\Big( \mathrm{logsumexp}_{i \in \mathcal{C}}\big(s_{t,i}^{(\ell,h)}\big) - \mathrm{logsumexp}_{i \leq t}\big(s_{t,i}^{(\ell,h)}\big)\Big),
\end{equation}
and the context bias $\mathbf{b}_t^{(\ell,h)} = \sum_{i \in \mathcal{C}} \alpha_{t,i}^{(\ell,h)} \, \mathbf{v}_i^{(\ell,h)}$.

For the fixed template tokens in the query block, we extract a set of operators across layers for each template position. For the query content tokens that vary across examples, we extract operators across content tokens and average them to obtain a single operator representing content positions in general. We also extract operators for the effect of ICL on the output tokens, which are computed on the tokens generated with the ICL prompt and averaged across output positions to form a unified generation operator. Our experiments demonstrate the effectiveness and importance of including the averaged operators for both content and generation positions to faithfully extract ICL knowledge.

To obtain a stable and reliable representation of the ICL effect induced by a fixed set of demonstrations, we compute the operator parameters on a set of $m$ validation examples without ground truth labels, and average these parameters across examples. While Equation~\eqref{eq:per-head-identity} holds exactly for each example, the averaged parameters provide a task-level approximation, which is supported by the cross-sample stability of the extracted parameters (Appendix~\ref{sec:appendix_stability_sample}). When multiple sets of demonstrations are available, we can further average the parameters across demonstration sets following \cite{huang2024multimodal} to obtain a more robust and task-level representation of the ICL knowledge.

\subsection{Knowledge Replay}
\label{sec:inference}

At test time, given a new query $x_q$, we construct a zero-shot prompt $P_{\mathrm{ZSL}}$ with the same query template as the ICL prompt $P_{\mathrm{ICL}}$ but without demonstrations. We then run the LLM on $P_{\mathrm{ZSL}}$ and inject the extracted operator parameters at the mapped positions in the query template and content tokens. At each mapped position $t'$, the original attention output is replaced with the values computed using the updated output projection layer:
\begin{equation}
\label{eq:hook}
\mathrm{Attn}_{t'}^{(\ell)} \;\leftarrow\; \widetilde{\mathbf{W}}_{O,t}^{(\ell)} \, \mathbf{U}_{t'}^{\mathrm{ZSL}} + \widetilde{\mathbf{b}}_{O,t}^{(\ell)},
\end{equation}

After the model steering during the prompt encoding, the LLM will then generate autoregressively with the task operator for generation positions installed at each decoding step, until the end of generation or a predefined maximum decoding length is reached.
Although the operator parameters are fixed at test time, the intervention remains input-dependent, as the scaling factors multiply the current zero-shot head outputs.
Since the operator only rescales and shifts the attention outputs, replay has essentially the same per-query cost as zero-shot inference (Appendix~\ref{sec:efficiency}).

\subsection{Knowledge Sparsification}
\label{sec:sparsification}

The full operator spans all layers and positions for model steering, but many sites may contribute negligibly to the model output. To understand how ICL knowledge is distributed across sites and to obtain an efficient, interpretable operator, we sparsify the full operator by intervening at each site and measuring its causal effect on the output distribution.

For each (layer, token) site $i$ and each of $m$ validation samples $v$, we compute the next-token negative log likelihood (NLL) on the teacher-forced ICL-generated tokens under three conditions: (1) with the full task operator, $\mathrm{NLL}_{\mathrm{full}}[v]$; (2) with all interventions disabled, $\mathrm{NLL}_{\mathrm{ZSL}}[v]$; and (3) with the operator applied everywhere except site $i$, which is patched to identity ($w \leftarrow 1$, $\mathbf{b} \leftarrow \mathbf{0}$ at site $i$ only), giving $\mathrm{NLL}_{\mathrm{perturb}}[v, i]$. The fraction of the operator's NLL gain that remains when site $i$ is perturbed defines the per-sample recovery:
\begin{equation}
\label{eq:recovery}
\mathrm{recovery}[v, i] = \frac{\mathrm{NLL}_{\mathrm{ZSL}}[v] - \mathrm{NLL}_{\mathrm{perturb}}[v, i]}{\mathrm{NLL}_{\mathrm{ZSL}}[v] - \mathrm{NLL}_{\mathrm{full}}[v]}.
\end{equation}

The per-sample importance is given by the loss in recovery, $\mathrm{importance}[v, i] = 1 - \mathrm{recovery}[v, i]$, and can be averaged across validation samples to provide a holistic view of how importance is distributed across sites. For robust site ranking, we use reciprocal rank fusion \cite{cormack2009reciprocal} to aggregate per-sample rankings, reducing sensitivity to NLL-scale variance across samples.
We use knowledge sparsification mainly for understanding the ICL mechanism, with additional performance results available in Appendix \ref{sec:sparsified_circuits}, where we show that the full operator remains a robust choice across tasks.

%% file: contents/experiments.tex
\section{Experiments}
\label{sec:experiments}

\subsection{Experimental Setup}
\label{sec:experimental_setup}

We evaluate four LLMs from different model families and sizes, including Qwen3-4B, Qwen3-8B \cite{yang2025qwen3}, Llama3.2-3B, and Llama3.1-8B \cite{grattafiori2024llama3}. Our evaluation covers eight tasks of varying difficulty, with two lexical tasks (Translation and Linguistic), three algorithmic tasks (Uppercase, Reverse, Deduplicate), and three reasoning tasks (GSM8K, MATH500, GPQA-Diamond). For lexical and algorithmic tasks, we use 128 held-out test samples for evaluation. For reasoning tasks, we rely on the standard test sets (GSM8K with 1319 samples, MATH500 with 500 samples, and GPQA-Diamond with 198 samples) and measure performance by exact match between the model's final answer and the reference answer. More details about the datasets are provided in Appendix~\ref{sec:appendix_dataset_details}.

We compare our proposed Task Operator (TO) with four ICL knowledge extraction baselines: Task Vectors (TV) \cite{hendel2023context}, Function Vectors (FV) \cite{todd2024function}, In-Context Vectors (ICV) \cite{liu2024incontext}, and Conceptors \cite{postmus2024steering}. For all methods, we use the same $K=8$ demonstrations for knowledge extraction and 32 answer-excluded validation samples for hyperparameter selection. We further compare with two trained input-dependent methods, ATV~\citep{kang2025adaptive} and ICR~\citep{li2026train}, in Section~\ref{sec:performance_comparison}. We also report the performance of zero-shot learning (ZSL) and standard 8-shot ICL as references to contextualize the results. All examples use the same prompt template, and for reasoning tasks, we append a ``Let's think step by step'' instruction to elicit chain-of-thought reasoning. Further implementation details for TO and the baselines are provided in Appendix~\ref{sec:appendix_implementation_details}. Additional experiments on the stability of extracted parameters, TO performance with sparsified circuits, robustness to prompt template choice, and cross-task transfer of extracted knowledge are available in Appendices \ref{sec:appendix_stability} - \ref{sec:knowledge_transfer}.

\subsection{Performance Comparison}
\label{sec:performance_comparison}

Table~\ref{tab:main_experiments} presents results across all models and tasks. Compared to other ICL knowledge extraction and replay methods, TO consistently achieves the highest accuracy on nearly every task-model combination, except in cases where ICL itself provides minimal or negative improvement over ZSL.

\input{tables/main_table}

On lexical tasks and simple algorithmic tasks like Uppercase, where prior methods such as TV already recover much of the ICL gain, TO matches or exceeds the best baseline, achieving near-ICL performance.
ICV is an exception that stays close to ZSL, even when its steering strength is tuned on the validation set for the lexical tasks (Appendix~\ref{sec:additional_baselines}), likely because it applies the same intervention at all positions, whereas Sections~\ref{sec:sparsification_analysis} and~\ref{sec:ablation_studies} show that different sites play distinct roles.
For more complex algorithmic tasks such as Reverse and Deduplicate, where output generation depends intricately on the input sequence, the baselines generally fail to capture and replay the ICL knowledge. This highlights their inherent limitations in representing the complex ICL knowledge required by these tasks. In contrast, TO effectively recovers a substantial portion of the ICL gain, showing its ability to capture the complex input-dependent knowledge induced by ICL.

On reasoning tasks, models already show strong zero-shot performance due to clear instructions and question formats, and ICL further improves accuracy in most cases. While the baselines improve over zero-shot, they remain far from ICL performance. In contrast, when ICL brings clear gains, as on the Qwen models, TO achieves a significant improvement over the baselines and recovers much of the ICL gain, demonstrating its ability to capture the complex reasoning patterns provided by the demonstrations.

Notably, TO does not simply outperform the baselines but also closely mimics the model's behavior under ICL. In task-model combinations where ICL provides only marginal improvement over zero-shot or even degrades performance, TO also fails to improve over ZSL. For example, on GPQA and MATH500 for both Llama models, where ICL gains are small or negative, TO shows correspondingly limited or no improvement, and in a few cases falls slightly below ZSL. This indicates that TO faithfully captures the internal mechanisms of ICL rather than providing a generic performance boost, reinforcing the validity of TO as a tool for understanding and replicating ICL behavior.

To assess the variability from sampling demonstrations, we repeat the comparison on Qwen3-4B and Qwen3-8B with multiple random seeds and a fixed test set (Table~\ref{tab:seed_robustness}).
Although accuracy varies with the sampled demonstrations, as is typical for ICL~\citep{zhang2022active,peng2024revisiting}, the advantage of TO is consistent. On Deduplicate, a paired comparison over the same seeds shows that TO outperforms FV ($p\approx0.01$) and TV ($p<0.01$) on both models, and on GPQA, TO exceeds all baselines by a clear margin.

\input{tables/seed_robustness}

Beyond Conceptor, a training-free input-dependent baseline, we compare TO with two trained input-dependent methods, ATV~\citep{kang2025adaptive} and ICR~\citep{li2026train}, each trained with $K \in \{8, 32, 128\}$ examples.
As shown in Table~\ref{tab:input_dependent}, ATV benefits from more training data, whereas ICR remains weak in this low-data regime.
In contrast, TO, which uses only eight demonstrations and no training, outperforms both methods in every setting except Qwen3-4B Deduplicate, where only ATV with $16\times$ more training examples surpasses it, suggesting that its analytically derived operator captures input-dependent knowledge that these trained methods need substantially more data to learn.

\input{tables/input_dependent}

\subsection{Sparse Circuit Analysis}
\label{sec:sparsification_analysis}

\begin{figure}[h!]
  \centering
  \includegraphics[width=\linewidth]{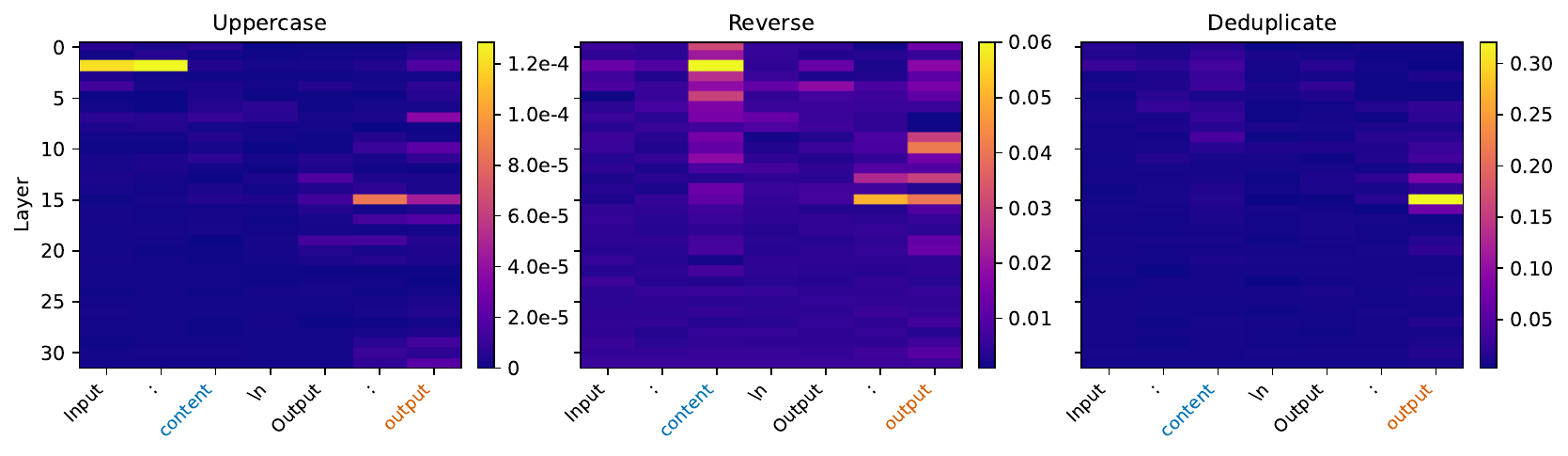}
  \caption{Per-site importance for three tasks on Llama3.1-8B. Brighter cells indicate higher importance. Token labels are color-coded by position type.}
  \label{fig:sparsity}
\end{figure}

We use the sparsification procedure from Section~\ref{sec:sparsification} to identify which sites are most important for the ICL behavior captured by TO. Figure~\ref{fig:sparsity} presents per-site importance maps for the algorithmic tasks on Llama3.1-8B, revealing both shared and task-specific patterns.

The importance maps show that the ICL knowledge captured by TO is often concentrated in a small subset of sites, with many sites having near-zero importance. However, the specific patterns vary by task, reflecting the different sparse circuits required for each. For the simpler Uppercase task, importance is highly localized, with most concentrated in template tokens within certain layers. In contrast, the more complex Reverse task displays a more distributed pattern, with importance spread across both template and content tokens, and some remaining at output positions. The Deduplicate task shows a focused pattern on output positions, especially in the middle layers.

These patterns help explain why baselines work reasonably well on Uppercase but fail on Reverse and Deduplicate. They mainly capture template-based ICL signals and miss the critical knowledge distributed across content and output positions in more complex tasks. Our ablation studies in Section~\ref{sec:ablation_studies} further validate these task-specific importance patterns, showing that knowledge in the identified areas contributes most to the improved performance over ZSL.

\subsection{Scaling Analysis}
\label{sec:scaling_analysis}

The extraction of ICL knowledge with TO involves two key hyperparameters: the number of demonstrations $K$ used in the prompt and the number of validation prompts $m$ averaged to estimate the per-site parameters. We analyze how varying each of these parameters affects the performance of TO on Llama3.1-8B across the three algorithmic tasks, comparing to the strong baseline, TV.

Table~\ref{tab:m_scaling} reports accuracy as $m$ increases with $K$ fixed at $8$. Overall, the performance of both TV and TO is largely recovered at $m{=}8$ and tends to converge as $m$ increases, while TO consistently outperforms TV at every $m$ on every task. Interestingly, increasing $m$ does not always increase the accuracy of TO, but it consistently pushes it toward the performance of standard ICL. For example, on Deduplicate, where TO performance at $m{=}8$ is lower than ICL, increasing $m$ to $32$ significantly improves TO and narrows the gap to ICL. However, on the Reverse task, where TO performance at $m{=}8$ is better than ICL, increasing $m$ to $32$ actually leads to a slight decrease in performance, making it closer to the ICL result. These results confirm the faithfulness of TO in replicating ICL behavior, and suggest that while a small number of validation prompts is sufficient to capture the ICL signal, increasing $m$ better aligns the replayed behavior with the original ICL performance.

\input{tables/m_scaling}

\begin{figure}[h!]
    \centering
    \includegraphics[width=0.9\linewidth]{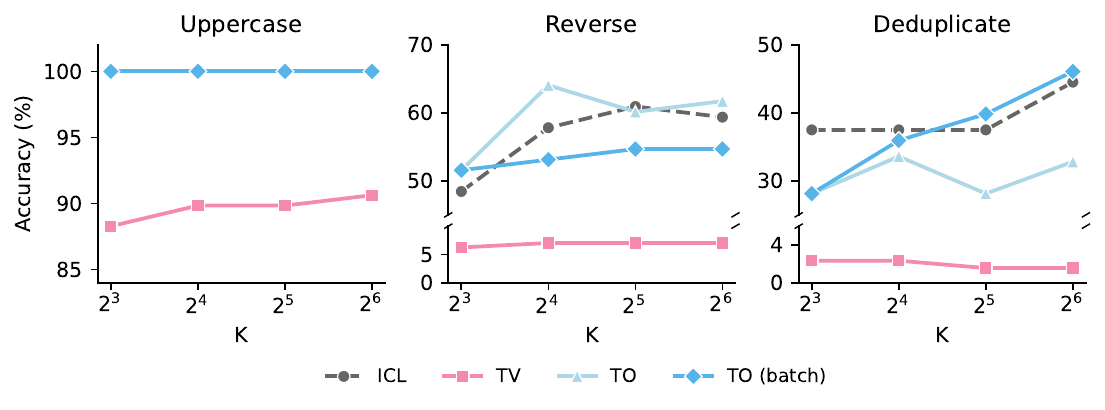}
    \caption{Effect of the number of demonstrations $K$ on test accuracy (Llama3.1-8B, $m{=}32$). TO (batch) averages operators from disjoint 8-shot batches.}
    \label{fig:k_scaling}
\end{figure}

Figure~\ref{fig:k_scaling} shows the accuracy of TV, TO, and standard ICL as $K$ increases with $m$ fixed at $32$. As expected, ICL performance generally improves with more demonstrations, though gains diminish and eventually plateau. TV shows only slight improvements as $K$ increases on Uppercase and Reverse, and remains far behind ICL. For TO, we observe an increase-then-decrease pattern on Reverse and Deduplicate as $K$ grows. We attribute this to $w_t^{(\ell,h)}$ in Equation~\eqref{eq:rescale} approaching zero for large $K$, which may lead to more sensitive and noisier replay behavior.

To address this and enable task operator extraction for many-shot learning, we break the large demonstration set into smaller batches (each with eight examples) and average the extracted operators across batches. This approach allows us to scale $K$ effectively without running into long-context bottlenecks, and produces more stable parameter estimates as the demonstration size is consistent across batches. We denote this batch-averaged version as TO (batch). As shown in Figure~\ref{fig:k_scaling}, TO (batch) shows clear non-decreasing performance with larger $K$ on all tasks, and may even outperform standard ICL in some settings (e.g., $K{=}64$ on Deduplicate). An alternative remedy that normalizes the context attention mass is discussed in Appendix~\ref{sec:nto}.

\subsection{Ablation Studies}
\label{sec:ablation_studies}

To understand the source of TO's superior performance over baselines, we conduct two ablation studies. Specifically, we ask whether TO's advantage comes from its ability to capture the ICL signal at specific positions missed by prior methods, or from its affine intervention structure that encodes dynamic, input-dependent knowledge beyond what fixed vectors can represent.

First, we assess the importance of different position types (template, content, and output) for capturing the ICL signal. We compare the template-only setting (used by baselines) to versions that also include content and output positions. This allows us to determine whether the ICL signal is primarily carried by the fixed instruction text, or also depends on sample-specific input and output tokens.

Second, we isolate the contribution of the multiplicative scaling component in the per-head operator by comparing the full affine operator to versions that use only the scaling or only the bias component. For a fair comparison, we re-optimize the parameters of the reduced operators in each condition. In the bias-only condition, we set $w_t^{(\ell,h)} = 1$ at every site and refit $\mathbf{b}_t^{(\ell,h)}$ as the mean difference between the full and context-masked head outputs, $\mathbf{u}_t^{(\ell,h)} - \mathbf{u}_t^{\prime(\ell,h)}$, across extraction samples. In the scaling-only condition, we set $\mathbf{b}_t^{(\ell,h)} = \mathbf{0}$ and refit $w_t^{(\ell,h)}$ via least-squares regression of $\mathbf{u}_t^{(\ell,h)}$ onto $\mathbf{u}_t^{\prime(\ell,h)}$. We use the template-only setting for analyzing the affine components, since template positions contain the fixed instruction text (identical across all test queries), making it easier to isolate the effect of the multiplicative component. In contrast, content and output positions carry sample-specific information, which naturally favors the affine operator for dynamic model steering.

Table~\ref{tab:position-type-ablation} presents the results of both ablations on Qwen3-4B, which most closely matches ICL performance among the tested models. The upper portion of the table shows the position-type ablation results, while the lower portion focuses on the affine component ablation at template positions. As shown, the full all-positions operator achieves the best overall performance across tasks. However, the importance of different position types varies substantially by task. For example, the simple Translation, Linguistic, and Uppercase tasks rely almost entirely on template positions, while Reverse, Deduplicate, and the reasoning tasks require significant contributions from content and output positions. Among the tasks where content and output positions matter, their relative importance also differs: Reverse relies more on content positions, while Deduplicate suffers a larger performance drop when output positions are removed. This suggests diverse ICL mechanisms across tasks and aligns with the importance patterns observed in Figure~\ref{fig:sparsity}.

\input{tables/ablation}

From the lower portion of Table~\ref{tab:position-type-ablation}, we see that the multiplicative scaling component is essential for capturing the ICL signal at template positions, while the additive bias alone is insufficient. This pattern holds across all tasks, and is especially pronounced on non-reasoning tasks, where both scaling-only and bias-only versions collapse to zero accuracy. These results indicate that the two components are individually insufficient but jointly effective, consistent with the affine structure analytically observed in the model's attention computations.

%% file: tables/main_table.tex
\begin{table}[h!] \small
\caption{Test accuracy (\%) across models and tasks. ICL denotes performance with $K=8$ examples provided. The scores without gray shading denote the performance with no in-context demonstrations.}
\centering
\label{tab:main_experiments}
\resizebox{\textwidth}{!}{
\begin{tabular}{clcccccccc}
\toprule
\multirow{2}{*}{Model} & \multirow{2}{*}{Method} &
\multicolumn{2}{c}{Lexical} &
\multicolumn{3}{c}{Algorithmic} &
\multicolumn{3}{c}{Reasoning} \\
\cmidrule(lr){3-4}\cmidrule(lr){5-7}\cmidrule(lr){8-10}
& & Translation & Linguistic & Uppercase & Reverse & Deduplicate & GSM8K & MATH500 & GPQA \\
\midrule
\multirow{7.5}{*}{\makecell{Qwen3\\(4B)}}
& \cellcolor{gray!15} ICL & \cellcolor{gray!15} 69.53 & \cellcolor{gray!15} 73.44 & \cellcolor{gray!15} 100.00 & \cellcolor{gray!15} 43.75 & \cellcolor{gray!15} 24.22 & \cellcolor{gray!15} 89.39 & \cellcolor{gray!15} 55.80 & \cellcolor{gray!15} 37.88 \\
\cmidrule(lr){2-10}
& ZSL & 0.00 & 0.00 & 0.00 & 0.00 & 0.00 & 74.53 & 23.60 & 17.68 \\
& TV & 43.75 & 49.22 & 92.19 & 1.56 & 0.00 & 81.05 & 28.00 & 21.21 \\
& FV & 0.00 & 17.97 & 94.53 & 0.00 & 16.41 & 73.16 & 26.00 & 15.15 \\
& ICV & 0.00 & 0.00 & 0.00 & 0.00 & 0.00 & 74.83 & 23.80 & 20.20 \\
& Conceptor & 38.28 & 50.78 & 96.09 & 0.78 & 0.78 & 79.83 & 26.20 & 16.16 \\
& TO (Ours) & \textbf{67.19} & \textbf{71.09} & \textbf{100.00} & \textbf{39.06} & \textbf{17.19} & \textbf{86.20} & \textbf{51.20} & \textbf{33.33} \\
\midrule
\multirow{7.5}{*}{\makecell{Qwen3\\(8B)}}
& \cellcolor{gray!15} ICL & \cellcolor{gray!15} 68.75 & \cellcolor{gray!15} 82.81 & \cellcolor{gray!15} 100.00 & \cellcolor{gray!15} 67.19 & \cellcolor{gray!15} 76.56 & \cellcolor{gray!15} 90.75 & \cellcolor{gray!15} 61.20 & \cellcolor{gray!15} 34.85 \\
\cmidrule(lr){2-10}
& ZSL & 0.00 & 0.00 & 0.00 & 0.00 & 0.00 & 58.76 & 19.00 & 21.21 \\
& TV & 49.22 & 64.84 & 94.53 & 9.38 & 9.38 & 73.69 & 26.00 & 22.73 \\
& FV & 36.72 & 53.91 & 98.44 & 21.09 & 57.03 & 63.61 & 22.40 & 19.70 \\
& ICV & 0.00 & 1.56 & 0.00 & 0.00 & 0.00 & 59.74 & 20.00 & 21.21 \\
& Conceptor & 49.22 & 60.94 & 91.41 & 4.69 & 4.69 & 70.74 & 25.40 & 18.18 \\
& TO (Ours) & \textbf{67.19} & \textbf{75.00} & \textbf{100.00} & \textbf{60.16} & \textbf{65.62} & \textbf{78.62} & \textbf{50.60} & \textbf{33.33} \\
\midrule
\multirow{7.5}{*}{\makecell{Llama3.2\\(3B)}}
& \cellcolor{gray!15} ICL & \cellcolor{gray!15} 66.41 & \cellcolor{gray!15} 75.78 & \cellcolor{gray!15} 100.00 & \cellcolor{gray!15} 21.88 & \cellcolor{gray!15} 33.59 & \cellcolor{gray!15} 69.98 & \cellcolor{gray!15} 27.20 & \cellcolor{gray!15} 24.24 \\
\cmidrule(lr){2-10}
& ZSL & 0.00 & 3.91 & 0.00 & 0.00 & 0.00 & 58.15 & 24.80 & 26.26 \\
& TV & 30.47 & 71.09 & 86.72 & 0.78 & 2.34 & 59.06 & \textbf{27.60} & 25.76 \\
& FV & 39.84 & 63.28 & 93.75 & 1.56 & 2.34 & 62.62 & 26.00 & 24.75 \\
& ICV & 0.00 & 4.69 & 0.00 & 0.00 & 0.00 & 58.00 & 24.80 & \textbf{30.81} \\
& Conceptor & 28.12 & 1.56 & 87.50 & 0.00 & 3.12 & 57.32 & 26.80 & 23.74 \\
& TO (Ours) & \textbf{67.97} & \textbf{75.78} & \textbf{100.00} & \textbf{31.25} & \textbf{24.22} & \textbf{67.78} & 25.20 & 22.73 \\
\midrule
\multirow{7.5}{*}{\makecell{Llama3.1\\(8B)}}
& \cellcolor{gray!15} ICL & \cellcolor{gray!15} 71.09 & \cellcolor{gray!15} 82.81 & \cellcolor{gray!15} 100.00 & \cellcolor{gray!15} 48.44 & \cellcolor{gray!15} 37.50 & \cellcolor{gray!15} 80.14 & \cellcolor{gray!15} 28.80 & \cellcolor{gray!15} 26.26 \\
\cmidrule(lr){2-10}
& ZSL & 1.56 & 10.94 & 0.00 & 0.00 & 0.00 & 37.91 & 27.60 & 23.74 \\
& TV & 53.12 & 75.00 & 88.28 & 6.25 & 2.34 & 41.32 & 26.00 & 20.71 \\
& FV & 64.06 & 63.28 & 50.78 & 6.25 & 2.34 & 41.70 & 28.20 & 23.23 \\
& ICV & 1.56 & 10.16 & 0.00 & 0.00 & 0.00 & 39.58 & 26.60 & \textbf{26.26} \\
& Conceptor & 47.66 & 41.41 & 82.03 & 7.03 & 2.34 & 38.67 & \textbf{31.80} & 25.25 \\
& TO (Ours) & \textbf{71.88} & \textbf{81.25} & \textbf{100.00} & \textbf{51.56} & \textbf{28.12} & \textbf{75.82} & 25.40 & 23.74 \\
\bottomrule
\end{tabular}%
}
\end{table}

%% file: tables/seed_robustness.tex
\begin{table}[h!]
\centering
\small
\caption{Accuracy (\%) with randomly sampled demonstrations, reported as mean $\pm$ standard deviation over 10 seeds for Deduplicate and 5 for GPQA.}
\label{tab:seed_robustness}
\begin{tabular}{lcccc}
\toprule
\multirow{2.5}{*}{Method} & \multicolumn{2}{c}{Qwen3 (4B)} & \multicolumn{2}{c}{Qwen3 (8B)} \\
\cmidrule(lr){2-3}\cmidrule(lr){4-5}
 & Deduplicate & GPQA & Deduplicate & GPQA \\
\midrule
ZSL & 0.0 & 17.7 & 0.0 & 21.2 \\
TV & 1.0$\pm$0.5 & 18.5$\pm$1.6 & 6.9$\pm$2.7 & 23.0$\pm$2.3 \\
FV & 16.6$\pm$5.1 & 17.5$\pm$1.8 & 55.8$\pm$5.8 & 20.3$\pm$1.2 \\
TO (Ours) & \textbf{26.9$\pm$9.8} & \textbf{29.1$\pm$3.5} & \textbf{63.9$\pm$8.7} & \textbf{29.5$\pm$2.7} \\
\bottomrule
\end{tabular}
\end{table}

%% file: tables/input_dependent.tex
\begin{table}[h!]
\centering
\small
\caption{Accuracy (\%) of trained input-dependent methods with $K$ training examples, compared with training-free TO using $K{=}8$ demonstrations.}
\label{tab:input_dependent}
\begin{tabular}{llccccccc}
\toprule
\multirow{2.5}{*}{Model} & \multirow{2.5}{*}{Task} & \multicolumn{3}{c}{ATV ($K$)} & \multicolumn{3}{c}{ICR ($K$)} & TO ($K$) \\
\cmidrule(lr){3-5}\cmidrule(lr){6-8}\cmidrule(lr){9-9}
 & & 8 & 32 & 128 & 8 & 32 & 128 & 8 \\
\midrule
\multirow{2}{*}{Qwen3 (4B)} & Deduplicate & 3.1 & 15.6 & \textbf{26.6} & 10.2 & 6.2 & 13.3 & 17.2 \\
 & Translation & 10.2 & 25.0 & 48.4 & 3.9 & 6.2 & 4.7 & \textbf{67.2} \\
\midrule
\multirow{2}{*}{Qwen3 (8B)} & Deduplicate & 56.2 & 42.2 & 60.9 & 32.0 & 42.2 & 42.2 & \textbf{65.6} \\
 & Translation & 50.8 & 60.2 & 64.8 & 2.3 & 4.7 & 5.5 & \textbf{67.2} \\
\bottomrule
\end{tabular}
\end{table}

%% file: tables/m_scaling.tex
\begin{table}[h] \small
\centering
\caption{Effect of the number of validation prompts $m$ on test accuracy (Llama3.1-8B, $K{=}8$ fixed). ICL ($K{=}8$) is shown as a reference baseline.}
\label{tab:m_scaling}
\begin{tabular}{llccccc}
\toprule
\textbf{Task} & \textbf{Method} & \textbf{$m=8$} & \textbf{$m=16$} & \textbf{$m=24$} & \textbf{$m=32$} & ICL \\
\midrule
\multirow{2}{*}{Uppercase}
 & TV & 87.50 & 88.28 & 87.50 & 88.28 & \multirow{2}{*}{100.00} \\
 & TO & 100.00 & 100.00 & 100.00 & 100.00 \\
\midrule
\multirow{2}{*}{Reverse}
 & TV & 5.47 & 6.25 & 6.25 & 6.25 & \multirow{2}{*}{48.44}\\
 & TO & 56.25 & 54.69 & 51.56 & 51.56 \\
\midrule
\multirow{2}{*}{Deduplicate}
 & TV & 2.34 & 2.34 & 2.34 & 2.34 & \multirow{2}{*}{37.50}\\
 & TO & 19.53 & 26.56 & 28.91 & 28.12 \\
\bottomrule
\end{tabular}
\end{table}

%% file: tables/ablation.tex
\begin{table}[h!] \small
\caption{Position-type and affine component ablations on Qwen3-4B. Top rows vary position scope. Bottom rows isolate the scaling ($w$) and bias ($\mathbf{b}$) components at template positions.}
\centering
\label{tab:position-type-ablation}
\resizebox{\textwidth}{!}{%
\begin{tabular}{lcccccccc}
\toprule
\multirow{2.5}{*}{Setting} &
\multicolumn{2}{c}{Lexical} &
\multicolumn{3}{c}{Algorithmic} &
\multicolumn{3}{c}{Reasoning} \\
\cmidrule(lr){2-3}\cmidrule(lr){4-6}\cmidrule(lr){7-9}
& Translation & Linguistic & Uppercase & Reverse & Deduplicate & GSM8K & MATH500 & GPQA \\
\midrule

\cellcolor{gray!15} ICL & \cellcolor{gray!15} 69.53 & \cellcolor{gray!15} 73.44 & \cellcolor{gray!15} 100.00 & \cellcolor{gray!15} 43.75 & \cellcolor{gray!15} 24.22 & \cellcolor{gray!15} 89.39 & \cellcolor{gray!15} 55.80 & \cellcolor{gray!15} 37.88 \\
\midrule

Template + Content + Output & \textbf{67.19} & 71.09 & \textbf{100.00} & \textbf{39.06} & \textbf{17.19} & \textbf{86.20} & 51.20 & \textbf{33.33} \\
Template + Output & \textbf{67.19} & 71.09 & \textbf{100.00} & 8.59 & 10.94 & 85.44 & \textbf{53.60} & 29.29 \\
Template + Content & 66.41 & 70.31 & \textbf{100.00} & 17.97 & 2.34 & 77.26 & 50.00 & 23.74 \\
Template ($w$ + $\mathbf{b}$) & \textbf{67.19} & \textbf{71.88} & \textbf{100.00} & 7.81 & 9.38 & 79.61 & 47.20 & 22.22 \\
Template ($w$ only) & 0.00 & 0.00 & 0.00 & 0.00 & 0.00 & 79.83 & 31.40 & 17.68 \\
Template ($\mathbf{b}$ only) & 0.00 & 0.00 & 0.00 & 0.00 & 0.00 & 34.87 & 10.20 & 19.19 \\

\bottomrule
\end{tabular}%
}
\end{table}

%% file: contents/conclusion.tex
\section{Conclusion}

We presented Task Operator (TO), a training-free method that captures in-context learning (ICL) knowledge as affine transformations of attention head outputs and replays them during zero-shot inference through dynamic updates to the output projection. Experiments across four language models and eight tasks demonstrate that TO achieves the best overall performance among prior methods, substantially closing the gap between zero-shot and few-shot performance. Our causal analysis reveals that the sites critical for ICL are sparse and task-specific, explaining why prior methods fall short on tasks requiring complex input-output interactions. Scaling experiments further show that averaging operators from disjoint demonstration batches enables effective many-shot learning without expanding the context window. These results connect in-context learning to an input-dependent reparameterization of the attention output projection, offering both a practical tool for efficient deployment and a lens for understanding how demonstrations reshape transformer computation.

%% file: contents/appendix.tex
\section{Limitations and Broader Impacts}
\label{sec:limitations}

\paragraph{Limitations.}
Task Operator (TO) provides a training-free way to extract and replay the effect of in-context demonstrations, while also opening several directions for further study. First, as shown in our scaling experiments, the performance of TO can become less stable when the number of demonstrations \(K\) is very large. This may occur because the attention mass assigned to non-context positions becomes small, which can increase the variance of the extracted parameters during replay. In practice, we find that averaging operators extracted from multiple independently sampled demonstration batches mitigates this issue and enables effective many-shot scaling. Second, our empirical evaluation focuses on text-only decoder language models across lexical, algorithmic, and reasoning tasks. These tasks cover a diverse set of input-output structures, but they do not exhaust the range of settings in which in-context learning is used. Extending TO to settings such as multimodal reasoning is a natural and promising direction for future work.

\paragraph{Broader Impacts.}
This work contributes to the mechanistic understanding of in-context learning by showing how the effect of demonstrations can be extracted from forward passes and replayed through TO. This perspective may help researchers analyze how transformer models encode task knowledge and why different tasks rely on different sparse circuits. TO may also make in-context learning more efficient in deployment by allowing demonstration-induced behavior to be reused without repeatedly processing the full demonstration context. This can reduce context-window usage and inference cost in settings where the same task is queried many times. As with other methods for adapting model behavior, extracted operators should be applied with appropriate validation and care, especially when demonstrations may contain sensitive, biased, or otherwise undesirable patterns.

\section{Dataset Details}
\label{sec:appendix_dataset_details}

We evaluate on eight tasks spanning three categories. All non-reasoning tasks use the template pair ``\texttt{Input: \{input\}\textbackslash{}nOutput: \{output\}}'' for demonstrations and ``\texttt{Input: \{input\}\textbackslash{}nOutput:}'' for queries. Reasoning tasks append ``\texttt{Let's think step by step.}'' to the output prefix, eliciting chain-of-thought (CoT) reasoning. For lexical and algorithmic tasks, 128 records form the held-out test set. Reasoning tasks use the original benchmark test splits.

\subsection{Lexical Tasks}

Both tasks use word-pair data released by \citet{dong2026understanding}.

\paragraph{Translation.}
English-to-French single-word translation.
The pool of 739 word pairs is allocated into 579 demonstration, 32 validation, and 128 test
records.

\begin{lstlisting}
Input: subject
Output: matière

Input: look
Output: voir

Input: wait
Output:
\end{lstlisting}
\textit{Expected output:}~\texttt{attendre}

\paragraph{Linguistic.}
English singular-to-plural inflection, covering regular and irregular forms.
The pool of 173 word pairs is split into 13 demonstration, 32 validation, and 128 test records.

\begin{lstlisting}
Input: bacterium
Output: bacteria

Input: seminoma
Output: seminomata

Input: foot
Output:
\end{lstlisting}
\textit{Expected output:}~\texttt{feet}

\subsection{Algorithmic Tasks}

All three tasks are generated synthetically over an eight-letter alphabet $\{a,\ldots,h\}$.
Each input is a random sequence of 5--12 space-separated tokens generated with a fixed seed.
Each split contains 1024 demonstration, 256 validation, and 128 test records.

\paragraph{Uppercase.}
Map each token in the sequence to its uppercase form.

\begin{lstlisting}
Input: e d g a g e d b
Output: E D G A G E D B

Input: b h a b a a g a c a g d
Output: B H A B A A G A C A G D

Input: d d f b b c b b
Output:
\end{lstlisting}
\textit{Expected output:}~\texttt{D D F B B C B B}

\paragraph{Reverse.}
Output the token sequence in reverse order.

\begin{lstlisting}
Input: d b f g d e
Output: e d g f b d

Input: g c h b d a
Output: a d b h c g

Input: b d a h f g d b h
Output:
\end{lstlisting}
\textit{Expected output:}~\texttt{h b d g f h a d b}

\paragraph{Deduplicate.}
Remove duplicate tokens, retaining only the first occurrence of each in order.

\begin{lstlisting}
Input: b h d e h
Output: b h d e

Input: f b f b d e
Output: f b d e

Input: g e h d h h
Output:
\end{lstlisting}
\textit{Expected output:}~\texttt{g e h d}

\subsection{Reasoning Tasks}

All three tasks use chain-of-thought (CoT) formatting.
The test split is the original benchmark's held-out set.
Demonstrations and validation records are sampled from the corresponding training split.

\paragraph{GSM8K \citep{cobbe2021training}.}
Grade-school arithmetic word problems with multi-step reasoning.
Intermediate computations are annotated as \texttt{<<expr=result>>}.
The 1024 demonstration and 256 validation records are sampled from the GSM8K training split.
The 1319 test records form the full official test set.
Targets end with \texttt{The answer is N}.

\begin{lstlisting}
Input: Elroy decides to enter a walk-a-thon [...] Last year, walkers earned
$4 a mile; this year $2.75. If last year's winner collected $44, how many
more miles will Elroy walk to collect the same amount?
Output: Let's think step by step. Last year's winner walked 11 miles because
44/4=<<44/4=11>>11. Elroy walks 16 miles because 44/2.75=<<44/2.75=16>>16.
Elroy walks 5 more miles because 16-11=<<16-11=5>>5. The answer is 5

Input: Sandy's monthly phone bill equals ten times her age. In two years she
will be three times Kim's age; Kim is 10. What is Sandy's phone bill?
Output: Let's think step by step. [...] The answer is 340

Input: Janet's ducks lay 16 eggs/day. She eats 3 for breakfast, bakes with 4,
and sells the rest at $2 each. How much does she make daily?
Output: Let's think step by step.
\end{lstlisting}
\textit{Expected output:}~\texttt{Janet sells 16-3-4=<<16-3-4=9>>9 eggs.
She makes 9*2=\$<<9*2=18>>18/day. The answer is 18}

\paragraph{MATH-500 \citep{lightman2023lets}.}
Competition-level mathematics problems (AMC/AIME style) from the Hendrycks MATH benchmark
\citep{hendrycks2021math}.
The 1024 demonstration and 256 validation records span seven subject areas from the Hendrycks
MATH training split. The 500 test records form the MATH-500 evaluation set.
Targets end with \texttt{The answer is X} where $X$ is the extracted boxed answer.

\begin{lstlisting}
Input: What is the smallest positive integer $x$ such that $400x$ is a
multiple of $576$?
Output: Let's think step by step. $400=2^4*5^2$, $576=2^6*3^2$. x must
contain $2^2*3^2=36$ as a factor. [...] The answer is 36

Input: Convert $3206_7$ to base 10.
Output: Let's think step by step. $3*7^3+2*7^2+0+6=1133$.
The answer is 1133

Input: Convert $(0,3)$ from rectangular to polar coordinates
(with $r>0$, $0\le\theta<2\pi$).
Output: Let's think step by step.
\end{lstlisting}
\textit{Expected output:}~\texttt{$r=3$, $\theta=\pi/2$. [...] The answer is $(3,\pi/2)$}

\paragraph{GPQA-Diamond \citep{rein2023gpqa}.}
Graduate-level multiple-choice questions in physics, chemistry, and biology.
The 198 test records are the GPQA-Diamond subset. The 218 demonstration and 32 validation
records are taken from GPQA-Main after removing Diamond questions.
Answer choices are shuffled per record with a fixed seed. Targets end with
\texttt{The answer is (LETTER)}.

\begin{lstlisting}
Input: methyllithium is added to dichlorotitanocene [...] What are the
multiplicities of the two most deshielded H nuclei in product 3?
(A) doublet, multiplet
(B) singlet, multiplet
(C) doublet, doublet
(D) doublet of doublets, doublet of doublets
Output: Let's think step by step. [...] The answer is (C)

Input: A transiting planet candidate is detected by TESS [...] Which
method CANNOT confirm its presence?
(A) Periodic FWHM changes in spectral lines
(B) Detection of a Rossiter-McLaughlin effect
(C) A signal in reflectance spectroscopy
(D) Periodic wavelength changes of spectral lines
Output: Let's think step by step. [...] The answer is (A)

Input: Two quantum states with energies E1 and E2 have a lifetime of
10^-9 sec and 10^-8 sec. Which energy difference allows clearly
resolving their spectral lines?
(A) 10^-4 eV
(B) 10^-11 eV
(C) 10^-9 eV
(D) 10^-8 eV
Output: Let's think step by step.
\end{lstlisting}
\textit{Expected output:}~\texttt{[\ldots] The energy difference must be >> 10\^{}-7 eV.
The answer is (A)}

\section{Implementation Details}
\label{sec:appendix_implementation_details}

\paragraph{Models.}
All experiments use instruction-tuned model variants.
Concretely, we use \texttt{Qwen/Qwen3-4B} and \texttt{Qwen/Qwen3-8B} from the Qwen3 family, and \texttt{meta-llama/Llama-3.2-3B-Instruct} and \texttt{meta-llama/Llama-3.1-8B-Instruct} from the Llama family, following their HuggingFace identifiers.

\paragraph{Task Operator extraction.}
The operator is extracted using $K{=}8$ demonstrations and $m{=}32$ answer-excluded validation prompts. In particular, we only keep validation records where ICL generation ends without hitting the token budget.
Each of the $T$ structural template token positions in the query receives an independent per-head $(w, \mathbf{b})$ pair, estimated by averaging across the validation prompts.
For content and output positions, the per-head parameters are first averaged over all token positions within that category in each prompt, then averaged across prompts, resulting in a single operator per (layer, head) that is applied uniformly to every content token and every generated token during zero-shot replay.
The all-positions, all-layers variant patches every decoder layer with no site pruning.

\paragraph{Baseline hyperparameters.}
All baselines use the same $K{=}8$ demonstrations and prompt templates as TO.
For TV \citep{hendel2023context}, FV \citep{todd2024function}, and Conceptor \citep{postmus2024steering}, the injection layer is selected by a full scan (step size~$1$) over all decoder layers, choosing the layer that minimizes NLL on the model's own ICL completions on the validation queries.
FV additionally identifies the top-$20$ attention heads globally across all layers by average indirect effect (AIE), computed as the reduction in zero-shot NLL when each head's pre-projection input is replaced with its mean ICL activation.
ICV \citep{liu2024incontext} injects the context vector at every layer simultaneously via a norm-preserving residual addition with scaling coefficient $\lambda{=}0.1$.
Conceptor uses aperture $\alpha{=}1.0$ and output scaling $\beta{=}1.0$.

\paragraph{Generation settings.}
All methods use a temperature of $0.0$ for deterministic decoding.
For lexical and algorithmic tasks, generation stops at the first newline with a token budget of $64$.
For reasoning tasks, generation stops at the substring \texttt{"\textbackslash{}nInput:"} (the ICL demo boundary) with a budget of $512$ tokens. The final answer is extracted as the text following the first occurrence of ``\texttt{The answer is}'' in the model's output.
The experiments are run on NVIDIA A100 GPUs with 80GB memory.

\section{Stability of Extracted Parameters}
\label{sec:appendix_stability}

The methodology aggregates per-prompt extractions in two ways.
First, at each fixed token position, we average across the $m$ ICL prompts (cross-sample).
Second, for tasks with variable-length content or output, the operator is applied at token positions not seen during extraction (cross-token within a category).
Each aggregation step relies on a distinct empirical stability claim, which we verify directly below.
We additionally include a robustness check on cross-demonstration-set stability.

\subsection{Cross-sample stability}
\label{sec:appendix_stability_sample}

For each (layer, head, token $j$), let $w_{j,i}^{(\ell,h)}$ denote the per-head scaling factor extracted from the $i$-th of $m{=}32$ ICL prompts.
Cross-sample stability is the claim that $\{w_{j,i}^{(\ell,h)}\}_{i=1}^m$ varies little across prompts at each fixed token position.
We measure this with the coefficient of variation
$\mathrm{CV}\big(w_j^{(\ell,h)}\big) = \mathrm{std}_i\big(w_{j,i}^{(\ell,h)}\big) / \big|\mathrm{mean}_i\big(w_{j,i}^{(\ell,h)}\big)\big|$,
computed independently at each token.
For the bias vector $\mathbf{b}_j^{(\ell,h)}$, we report the average pairwise cosine similarity across the $m$ prompts at each token position.

We extract per-prompt operators using all position types and all decoder layers (the same all-positions, all-layers configuration as in the main experiments) and run $m{=}32$ ICL prompts per (model, task) cell.
We test the four LLMs on one representative task from each category: Translation for lexical, Reverse for algorithmic, and GSM8K for reasoning.
At each (layer, token, head) site, we record one $\mathrm{CV}(w)$ value across the $m$ prompts and one mean pairwise $\cos(\mathbf{b})$ value across the $\binom{m}{2}$ off-diagonal pairs of bias vectors.

Table~\ref{tab:cross_sample_stability} reports, per (model, task) cell, the fraction of sites whose per-prompt $\mathrm{CV}(w)$ falls below $\{0.1, 0.2, 0.5\}$ and whose mean pairwise $\cos(\mathbf{b})$ exceeds $\{0.9, 0.8, 0.5\}$.
Averaged across the twelve cells, $80.7\%$ of sites have $\mathrm{CV}(w) {<} 0.1$ and $92.1\%$ have $\cos(\mathbf{b}) {>} 0.9$.
The Llama models are noticeably tighter than the Qwen models on both metrics, with the fraction at $\mathrm{CV}(w) {<} 0.1$ shifting from $70$--$77\%$ on the Qwen models to $85$--$92\%$ on the Llama models.
Within each model, Reverse is the tightest task and GSM8K consistently exhibits the lowest fraction of sites at $\cos(\mathbf{b}) {>} 0.9$.
The per-prompt scaling factors and bias directions cluster tightly at each fixed token position, supporting cross-sample averaging as the basis for the broadcast operator and the cross-sample stability claim invoked in Section~\ref{sec:extraction}.

\begin{table}[h]
\centering
\caption{Cross-sample stability across ICL prompts. Each row reports the fraction of (layer, token, head) sites (in \%) whose per-prompt scaling factor satisfies $\mathrm{CV}(w)$ below the threshold and whose per-prompt bias satisfies mean pairwise $\cos(\mathbf{b})$ above the threshold.}
\label{tab:cross_sample_stability}
\small
\setlength{\tabcolsep}{4pt}
\begin{tabular}{llcccccc}
\toprule
\multirow{2.5}{*}{Model} & \multirow{2.5}{*}{Task} & \multicolumn{3}{c}{$\mathrm{CV}(w) {<}$} & \multicolumn{3}{c}{$\cos(\mathbf{b}) {>}$} \\
\cmidrule(lr){3-5}\cmidrule(lr){6-8}
& & $0.1$ & $0.2$ & $0.5$ & $0.9$ & $0.8$ & $0.5$ \\
\midrule
\multirow{3}{*}{Qwen3-4B} & Translation & 70.4 & 83.9 & 95.4 & 88.1 & 98.0 & 100.0 \\
                          & Reverse     & 77.3 & 90.0 & 97.1 & 97.7 & 99.6 & 100.0 \\
                          & GSM8K       & 71.4 & 85.1 & 94.3 & 83.5 & 92.7 &  99.2 \\
\midrule
\multirow{3}{*}{Qwen3-8B} & Translation & 71.9 & 84.9 & 96.0 & 86.2 & 97.4 & 100.0 \\
                          & Reverse     & 76.6 & 89.5 & 96.3 & 97.8 & 99.7 & 100.0 \\
                          & GSM8K       & 73.1 & 85.9 & 94.7 & 83.9 & 93.2 &  99.3 \\
\midrule
\multirow{3}{*}{Llama3.2-3B} & Translation & 89.3 & 96.7 & 99.4 & 95.5 & 99.3 & 100.0 \\
                                       & Reverse     & 92.0 & 97.7 & 99.7 & 98.7 & 99.9 & 100.0 \\
                                       & GSM8K       & 87.0 & 95.6 & 99.2 & 93.1 & 97.9 & 100.0 \\
\midrule
\multirow{3}{*}{Llama3.1-8B} & Translation & 85.3 & 94.7 & 99.2 & 93.1 & 99.0 & 100.0 \\
                                       & Reverse     & 88.8 & 96.0 & 99.3 & 97.9 & 99.8 & 100.0 \\
                                       & GSM8K       & 85.3 & 94.7 & 99.3 & 89.8 & 96.0 &  99.8 \\
\bottomrule
\end{tabular}
\end{table}

We further examine how stability and replay accuracy evolve with the number of extraction samples $m$ on Llama3.1-8B Deduplicate (Table~\ref{tab:stability_m}).
The extracted parameters are already highly stable at $m{=}8$, and both stability metrics remain nearly unchanged as $m$ increases, while accuracy improves up to $m{=}24$ and then plateaus, consistent with Table~\ref{tab:m_scaling}.
This suggests that additional extraction samples mainly refine the estimates at a small subset of sites that are critical for performance.

\begin{table}[h]
\centering
\caption{Cross-sample stability and TO accuracy as the number of extraction samples $m$ increases (Llama3.1-8B, Deduplicate).}
\label{tab:stability_m}
\small
\begin{tabular}{lcccc}
\toprule
 & $m{=}8$ & $m{=}16$ & $m{=}24$ & $m{=}32$ \\
\midrule
Sites with $\mathrm{CV}(w) {<} 0.1$ (\%) & 88.9 & 89.7 & 89.3 & 89.1 \\
Sites with $\cos(\mathbf{b}) {>} 0.9$ (\%) & 98.4 & 99.1 & 99.0 & 99.0 \\
TO accuracy (\%) & 19.5 & 26.6 & 28.9 & 28.1 \\
\bottomrule
\end{tabular}
\end{table}

\subsection{Cross-token stability within a category}
\label{sec:appendix_stability_token}

For content and output categories, our method averages the extracted $(w, \mathbf{b})$ across all token offsets within each category, producing a single broadcast operator applied uniformly regardless of position.
We assess whether this averaging discards useful offset-specific structure by comparing the broadcast variant against a per-rank variant that stores independent $(w, \mathbf{b})$ for each of the first $12$ token offsets in the content and output categories.
We use $12$ offsets because this provides complete coverage for Translation (single-word content and output) and Reverse (sequences of 5--12 tokens), making the comparison maximally informative on these tasks: any offset-specific structure that exists will be captured by the per-rank variant.
For GSM8K, where content and output sequences span dozens to hundreds of tokens, exhaustive per-rank coverage is infeasible. In this case, $12$ offsets covers a prefix.

Table~\ref{tab:cross_token_stability} reports the accuracy of each variant on $n{=}128$ test records per (model, task) cell, using the same four models and three tasks as in Section~\ref{sec:appendix_stability_sample}.
On Translation, the two variants match exactly ($\Delta = 0$) for all models, confirming that no exploitable offset-specific structure exists when content spans are short.
On Reverse, where per-rank also has complete offset coverage, broadcast still outperforms by $2$--$10$\,pp.
This shows that the averaged operator is more robust than offset-specific parameters even with full positional coverage, likely because averaging pools signal across the $m$ validation prompts at each (layer, head) site, whereas per-rank must estimate $12{\times}$ as many parameters from the same data, adding additional noises.
On GSM8K, broadcast matches or outperforms per-rank in three of four models (gaps of $-5$ to $-28$\,pp), with a single marginal exception (Llama3.1-8B, $+2.3$\,pp).
Here the long sequences make exhaustive per-rank coverage infeasible, but broadcast remains effective by capturing the shared signal across all offsets without positional alignment.
Across all settings, the broadcast variant matches or exceeds per-rank, confirming that averaging within a category is not only sufficient but preferable to offset-specific modeling.

\begin{table}[h]
\centering
\caption{Cross-token stability via downstream accuracy. The per-rank variant uses offset-aware $(w, \mathbf{b})$ for the first $12$ offsets in content and output; broadcast averages across offsets to obtain one $(w, \mathbf{b})$ per category. $\Delta$: per-rank minus broadcast accuracy.}
\label{tab:cross_token_stability}
\small
\begin{tabular}{llccc}
\toprule
Model & Task & Broadcast (\%) & Per-rank (\%) & $\Delta$ (pp) \\
\midrule
\multirow{3}{*}{Qwen3-4B} & Translation & 67.2 & 67.2 & $\phantom{-}0.0$ \\
                          & Reverse     & 39.1 & 36.7 & $-2.3$ \\
                          & GSM8K       & 86.2 & 81.2 & $-5.0$ \\
\midrule
\multirow{3}{*}{Qwen3-8B} & Translation & 67.2 & 67.2 & $\phantom{-}0.0$ \\
                          & Reverse     & 60.2 & 50.0 & $-10.2$ \\
                          & GSM8K       & 78.6 & 50.8 & $-27.8$ \\
\midrule
\multirow{3}{*}{Llama3.2-3B} & Translation & 68.0 & 68.0 & $\phantom{-}0.0$ \\
                                       & Reverse     & 31.2 & 21.1 & $-10.2$ \\
                                       & GSM8K       & 67.8 & 60.9 & $-6.8$ \\
\midrule
\multirow{3}{*}{Llama3.1-8B} & Translation & 71.9 & 71.9 & $\phantom{-}0.0$ \\
                                       & Reverse     & 51.6 & 47.7 & $-3.9$ \\
                                       & GSM8K       & 75.8 & 78.1 & $\phantom{-}+2.3$ \\
\bottomrule
\end{tabular}
\end{table}

\subsection{Cross-demonstration-set stability}
\label{sec:appendix_stability_demoset}

Beyond the two main premises, we check whether the extracted operator is genuinely task-specific rather than demonstration-set-specific.
For a given task, we extract two operators $(w, \mathbf{b})^{(1)}$ and $(w, \mathbf{b})^{(2)}$ from disjoint pools of $m$ prompts using different demonstration sets, and compare their per-slot agreement.

We instantiate the two operators using non-overlapping demonstration pools sampled from the same task.
Operator $(w, \mathbf{b})^{(1)}$ uses the first eight demonstrations and the first $32$ validation queries.
Operator $(w, \mathbf{b})^{(2)}$ uses the next eight demonstrations (indices $8$ through $15$) with the same validation queries.
The two pools share no demonstration prompts, so any agreement between the two operators at a fixed (layer, token, head) site reflects the task signal rather than the identity of the eight demonstrations.

We compute two per-slot agreement metrics.
Pearson correlation $\rho(w^{(1)}, w^{(2)})$ is taken over the flattened vector of per-slot scaling factors across all (layer, token, head) sites in a cell, resulting in one scalar per (model, task).
Cosine similarity $\cos(\mathbf{b}^{(1)}, \mathbf{b}^{(2)})$ is computed at each site separately, and we report its mean and median across sites within each cell.

Table~\ref{tab:cross_demoset_stability} reports the agreement statistics across the same four models and three tasks.
Pearson correlations of $w$ exceed $0.96$ in every cell, with a mean of $0.984$ across the twelve cells, indicating that the per-slot scaling pattern is essentially fixed by the task.
Mean cosine similarity of $\mathbf{b}$ falls in the range $0.81$--$0.95$, with Reverse showing the highest agreement and Translation the lowest.
Together, these statistics support the claim that the extracted operator captures the task-specific signal rather than the identity of the demonstrations used to extract it.

\begin{table}[h]
\centering
\caption{Cross-demonstration-set stability. Pearson($w$) is computed once per (model, task) cell over all flattened (layer, token, head) sites. Cosine similarity of $\mathbf{b}$ is computed per site, with mean and median aggregated across sites within each cell.}
\label{tab:cross_demoset_stability}
\small
\setlength{\tabcolsep}{4pt}
\begin{tabular}{llccc}
\toprule
Model & Task & Pearson($w$) & Mean $\cos(\mathbf{b})$ & Median $\cos(\mathbf{b})$ \\
\midrule
\multirow{3}{*}{Qwen3-4B} & Translation & 0.981 & 0.844 & 0.896 \\
                          & Reverse     & 0.992 & 0.953 & 0.967 \\
                          & GSM8K       & 0.993 & 0.912 & 0.947 \\
\midrule
\multirow{3}{*}{Qwen3-8B} & Translation & 0.982 & 0.835 & 0.891 \\
                          & Reverse     & 0.992 & 0.954 & 0.968 \\
                          & GSM8K       & 0.993 & 0.908 & 0.945 \\
\midrule
\multirow{3}{*}{Llama3.2-3B} & Translation & 0.974 & 0.813 & 0.864 \\
                                       & Reverse     & 0.987 & 0.941 & 0.962 \\
                                       & GSM8K       & 0.980 & 0.899 & 0.933 \\
\midrule
\multirow{3}{*}{Llama3.1-8B} & Translation & 0.964 & 0.839 & 0.892 \\
                                       & Reverse     & 0.982 & 0.950 & 0.963 \\
                                       & GSM8K       & 0.991 & 0.911 & 0.933 \\
\bottomrule
\end{tabular}
\end{table}

\section{Task Operator Performance with Sparsified Circuits}
\label{sec:sparsified_circuits}

\begin{figure}[h!]
  \centering
  \includegraphics[width=\linewidth]{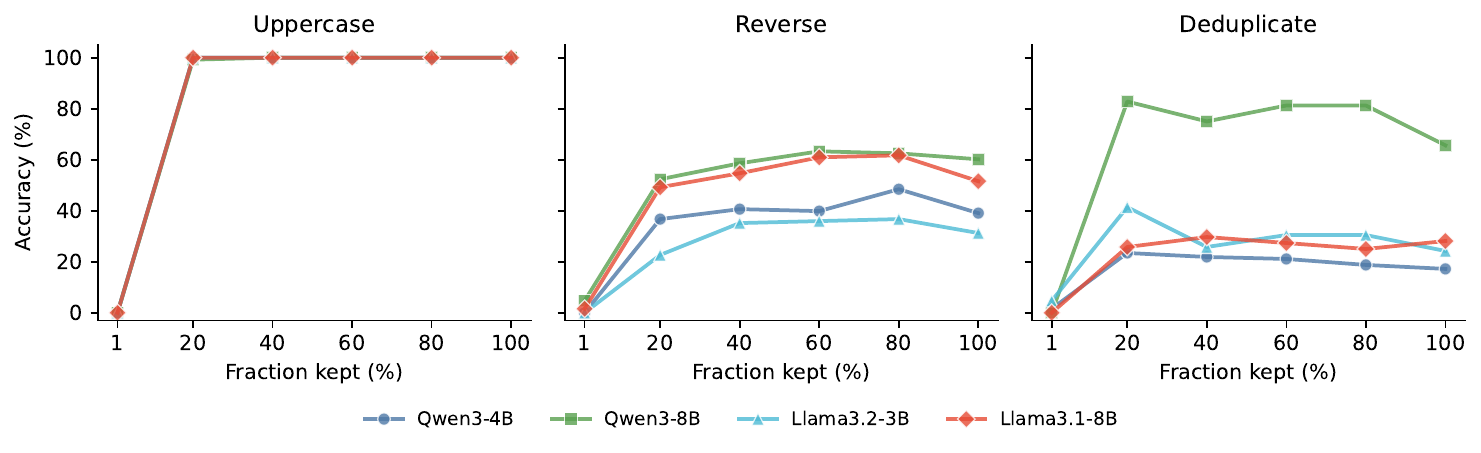}
  \caption{Test accuracy under top-$p$ sparsification on three algorithmic tasks across four models.}
  \label{fig:sparsity_ablation}
\end{figure}

We next evaluate whether the sparse circuits identified by the validation-NLL criterion are sufficient for task-operator replay. Starting from the dense task
operator, we rank intervention sites using the site-importance procedure in Section \ref{sec:sparsification} and retain only the top-$p$ fraction of sites. Figure~\ref{fig:sparsity_ablation} reports the resulting test accuracy as $p$ is swept from 1\% to 100\% on the three algorithmic tasks across all four models.

The sweep reveals two main trends. First, retaining only the top 1\% of sites is generally insufficient: performance collapses to near-zero on almost all model--task pairs. Thus, although the ICL signal is sparse, successful replay does not reduce to a single dominant site. Instead, it requires a small but coordinated set of interventions. Second, moderate sparsification preserves most of the dense operator's performance. In particular, retaining roughly 20\% of sites already recovers nearly the full dense-operator accuracy on Uppercase and Reverse, indicating that much of the useful ICL signal is concentrated in a small subset of layer-token positions. On Deduplicate, sparsification can be even more beneficial. For Qwen3-8B, accuracy increases from 65.6\% with the dense operator to 82.8\% at $p=20\%$, suggesting that low-importance sites can introduce interference rather than useful task information.

To choose a sparsification level without using test labels, we select $p^\star$ separately for each model--task pair by minimizing teacher-forced validation NLL over $p \in \{0.2, 0.4, 0.6, 0.8, 1.0\}$, breaking ties in favor of smaller circuits. Site rankings are obtained by reciprocal-rank fusion over the per-sample NLL impacts. Table~\ref{tab:sparsified_task_operator} compares the resulting sparse operator against the dense operator. The validation-selected sparse operator preserves or improves test accuracy in all 12 model--task cells while retaining, on average, 67\% of sites. The largest gains appear on Reverse and Deduplicate, where pruning low-importance sites reduces interference and improves accuracy by up to 9.4 percentage points. These results support the interpretation that task operators are not only sparse in their causal structure, but can also be made more effective by removing sites that contribute little or noisy signal. 

However, selecting the optimal sparsification level requires sweeping multiple thresholds on the validation set, which introduces an additional hyperparameter selection step and may overfit to validation-specific noise. 
For example, Table~\ref{tab:sparsified_task_operator} shows that the optimal $p^\star$ is selected as 100\% for some models on the Uppercase task. However, Figure~\ref{fig:sparsity_ablation} reflects the sparse nature of the Uppercase task, indicating that the $p^\star$ might overfit the validation samples.
For this reason, we use the full operator, corresponding to $p=100\%$, in the main experiments. This choice avoids threshold tuning while still providing strong and stable performance, as Figure~\ref{fig:sparsity_ablation} shows that the dense operator remains competitive across tasks and models. We therefore view sparsification primarily as an analysis tool for identifying compact task circuits, rather than as a necessary component of the main TO method.

\begin{table}[h]
\centering
\caption{Test accuracy (\%) of the dense and val-NLL-sparsified task operator on the three algorithmic tasks. $p^\star$ is selected by minimizing validation NLL over $p \in \{0.2, 0.4, 0.6, 0.8, 1.0\}$. \emph{Sites kept} is the fraction of sites retained at $p^\star$.}
\label{tab:sparsified_task_operator}
\small
\begin{tabular}{llccc}
\toprule
Model & Method & Uppercase & Reverse & Deduplicate \\
\midrule

\multirow{4.5}{*}{\makecell{Qwen3\\(4B)}}
& \cellcolor{gray!15} ICL & \cellcolor{gray!15} 100.00 & \cellcolor{gray!15} 43.75 & \cellcolor{gray!15} 24.22 \\
\cmidrule(lr){2-5}
& TO (Full) & 100.00 & 39.06 & 17.19 \\
& TO (Sparse @ $p^\star$) & \textbf{100.00} & \textbf{39.84} & \textbf{21.09} \\
& \textit{Sites kept (\%)} & 100 & 60 & 60 \\
\midrule
\multirow{4.5}{*}{\makecell{Qwen3\\(8B)}}
& \cellcolor{gray!15} ICL & \cellcolor{gray!15} 100.00 & \cellcolor{gray!15} 67.19 & \cellcolor{gray!15} 76.56 \\
\cmidrule(lr){2-5}
& TO (Full) & 100.00 & 60.16 & 65.62 \\
& TO (Sparse @ $p^\star$) & \textbf{100.00} & \textbf{63.28} & \textbf{75.00} \\
& \textit{Sites kept (\%)} & 100 & 60 & 40 \\
\midrule
\multirow{4.5}{*}{\makecell{Llama3.2\\(3B)}}
& \cellcolor{gray!15} ICL & \cellcolor{gray!15} 100.00 & \cellcolor{gray!15} 21.88 & \cellcolor{gray!15} 33.59 \\
\cmidrule(lr){2-5}
& TO (Full) & 100.00 & 31.25 & 24.22 \\
& TO (Sparse @ $p^\star$) & \textbf{100.00} & \textbf{36.72} & \textbf{30.47} \\
& \textit{Sites kept (\%)} & 60 & 80 & 60 \\
\midrule
\multirow{4.5}{*}{\makecell{Llama3.1\\(8B)}}
& \cellcolor{gray!15} ICL & \cellcolor{gray!15} 100.00 & \cellcolor{gray!15} 48.44 & \cellcolor{gray!15} 37.50 \\
\cmidrule(lr){2-5}
& TO (Full) & 100.00 & 51.56 & 28.12 \\
& TO (Sparse @ $p^\star$) & \textbf{100.00} & \textbf{60.94} & \textbf{29.69} \\
& \textit{Sites kept (\%)} & 80 & 60 & 40 \\
\bottomrule
\end{tabular}
\end{table}

\section{Robustness to Prompt Template Choice}

The main experiments use the canonical \texttt{Input}/\texttt{Output} prompt format. Since ICL and activation-based replay methods can be sensitive to prompt wording, we further evaluate whether TO remains effective when the same task is expressed with different surface templates.
We consider three templates. Template \texttt{t0} is the canonical format with \texttt{Input} and \texttt{Output} field names. Template \texttt{t1} uses the corresponding \texttt{Q} and \texttt{A} field names: ``\texttt{Q: \{input\}\textbackslash{}nA: \{output\}}''. Template \texttt{t2} uses a compact arrow format: ``\texttt{\{input\} -> \{output\}}''. For each model, task, and template, TO and TV are extracted and evaluated under the same template.

Table~\ref{tab:template_robustness} reports results on one representative task from each category, with Translation for lexical tasks, Deduplicate for algorithmic tasks, and GPQA for reasoning. Across these settings, TO remains the stronger intervention method in most cases and is substantially more stable than TV on tasks where the prompt format strongly affects the baseline. On Translation, TO stays close to ICL across all three templates, while TV is more sensitive to template choice. On Deduplicate, performance varies with the prompt format for both ICL-style prompting and replay methods, but TO consistently remains well above TV, indicating that the operator continues to capture useful task structure under matched template changes. On GPQA, where zero-shot performance is already competitive and the margins between non-ICL methods are smaller, TO remains broadly comparable across templates.

Overall, these results suggest that TO is not tied to a single prompt wording when extraction and evaluation use the same template. TO preserves its advantage over the task vector baseline on the lexical and algorithmic tasks and remains competitive on the reasoning task. This behavior is consistent with TO's design. Instead of relying on a single fixed residual vector, TO applies affine operators across layers, attention heads, and position types, making the replayed ICL effect less dependent on any one prompt token or extraction site.

\begin{table}[h]
\centering
\caption{
Test accuracy (\%) under three matched prompt templates. \texttt{t0} denotes the
canonical \texttt{Input}/\texttt{Output} format, \texttt{t1} the \texttt{Q}/\texttt{A}
format, and \texttt{t2} the compact arrow format. TO and TV are extracted and evaluated
under the same template. Bold marks the best score among ZSL, TV, and TO.
}
\label{tab:template_robustness}
\resizebox{\textwidth}{!}{%
\begin{tabular}{llccccccccc}
\toprule
\multirow{2}{*}{Model} & \multirow{2}{*}{Method} &
\multicolumn{3}{c}{Translation} &
\multicolumn{3}{c}{Deduplicate} &
\multicolumn{3}{c}{GPQA} \\
\cmidrule(lr){3-5}\cmidrule(lr){6-8}\cmidrule(lr){9-11}
& & $t_0$ & $t_1$ & $t_2$ & $t_0$ & $t_1$ & $t_2$ & $t_0$ & $t_1$ & $t_2$ \\
\midrule
\multirow{4}{*}{\makecell{Qwen3\\(4B)}}
& \cellcolor{gray!15} ICL & \cellcolor{gray!15} 69.53 & \cellcolor{gray!15} 67.97 & \cellcolor{gray!15} 67.19 & \cellcolor{gray!15} 24.22 & \cellcolor{gray!15} 14.84 & \cellcolor{gray!15} 18.75 & \cellcolor{gray!15} 37.88 & \cellcolor{gray!15} 29.80 & \cellcolor{gray!15} 29.29 \\
\cmidrule(lr){2-11}
& ZSL        &  0.00 &  0.00 &  0.00 &  0.00 &  0.00 &  0.00 & 17.68 & 16.16 & 17.68 \\
& TV         & 43.75 & 41.41 & 11.72 &  0.00 &  0.78 &  0.78 & 21.21 & \textbf{30.30} & 23.23 \\
& TO (Ours)  & \textbf{67.19} & \textbf{64.84} & \textbf{62.50} & \textbf{17.19} & \textbf{15.62} & \textbf{ 9.38} & \textbf{33.33} & 29.29 & \textbf{28.79} \\
\midrule
\multirow{4}{*}{\makecell{Qwen3\\(8B)}}
& \cellcolor{gray!15} ICL & \cellcolor{gray!15} 68.75 & \cellcolor{gray!15} 67.97 & \cellcolor{gray!15} 65.62 & \cellcolor{gray!15} 76.56 & \cellcolor{gray!15} 47.66 & \cellcolor{gray!15} 35.16 & \cellcolor{gray!15} 34.85 & \cellcolor{gray!15} 32.83 & \cellcolor{gray!15} 34.85 \\
\cmidrule(lr){2-11}
& ZSL        &  0.00 &  0.00 &  0.00 &  0.00 &  0.00 &  0.00 & 21.21 & 23.23 & 18.69 \\
& TV         & 49.22 & 32.81 &  4.69 &  9.38 &  1.56 &  0.00 & 22.73 & 23.23 & 28.79 \\
& TO (Ours)  & \textbf{67.19} & \textbf{67.19} & \textbf{62.50} & \textbf{65.62} & \textbf{36.72} & \textbf{ 7.81} & \textbf{33.33} & \textbf{32.83} & \textbf{29.29} \\
\midrule
\multirow{4}{*}{\makecell{Llama3.2\\(3B)}}
& \cellcolor{gray!15} ICL & \cellcolor{gray!15} 66.41 & \cellcolor{gray!15} 66.41 & \cellcolor{gray!15} 61.72 & \cellcolor{gray!15} 33.59 & \cellcolor{gray!15} 21.09 & \cellcolor{gray!15} 20.31 & \cellcolor{gray!15} 24.24 & \cellcolor{gray!15} 25.25 & \cellcolor{gray!15} 22.73 \\
\cmidrule(lr){2-11}
& ZSL        &  0.00 &  0.78 &  3.91 &  0.00 &  0.00 &  1.56 & \textbf{26.26} & 28.28 & 22.73 \\
& TV         & 30.47 & 48.44 & 53.91 &  2.34 &  0.00 &  5.47 & 25.76 & 24.75 & \textbf{28.28} \\
& TO (Ours)  & \textbf{67.97} & \textbf{67.19} & \textbf{66.41} & \textbf{24.22} & \textbf{ 6.25} & \textbf{ 8.59} & 22.73 & \textbf{28.79} & 26.77 \\
\midrule
\multirow{4}{*}{\makecell{Llama3.1\\(8B)}}
& \cellcolor{gray!15} ICL & \cellcolor{gray!15} 71.09 & \cellcolor{gray!15} 67.97 & \cellcolor{gray!15} 65.62 & \cellcolor{gray!15} 37.50 & \cellcolor{gray!15} 35.94 & \cellcolor{gray!15} 28.91 & \cellcolor{gray!15} 26.26 & \cellcolor{gray!15} 24.75 & \cellcolor{gray!15} 23.74 \\
\cmidrule(lr){2-11}
& ZSL        &  1.56 &  3.12 &  3.91 &  0.00 &  0.00 &  2.34 & \textbf{23.74} & \textbf{31.82} & 19.19 \\
& TV         & 53.12 & 66.41 & 64.06 &  2.34 &  1.56 &  6.25 & 20.71 & 22.73 & \textbf{24.75} \\
& TO (Ours)  & \textbf{71.88} & \textbf{69.53} & \textbf{67.97} & \textbf{28.12} & \textbf{28.91} & \textbf{18.75} & \textbf{23.74} & 27.27 & 23.74 \\
\bottomrule
\end{tabular}
}
\end{table}

\section{Cross-Task Transfer of Extracted Knowledge}
\label{sec:knowledge_transfer}

The stability analyses in Section \ref{sec:appendix_stability} show that task operators are not artifacts of individual prompts or demonstration sets. We next ask whether the extracted knowledge can generalize across related reasoning tasks. We extract an operator from GSM8K using the same setting as the main experiments, with $K=8$ demonstrations and $m=32$ validation prompts, and replay it on MATH-500 and GPQA-Diamond. During transfer, the input prompt remains target specific, while only the source of the operator changes. We compare this setting with zero-shot inference, standard in-domain task operators, and full in-context learning on the target task.

Table~\ref{tab:transfer} shows that the GSM8K operator transfers strongly to MATH-500. Across model families, its performance remains close to the in-domain operator, and in some cases it is competitive with or better than full in-context learning. This result suggests that TO captures a reusable chain-of-thought reasoning component rather than merely memorizing task-specific demonstrations.

Transfer to GPQA-Diamond is more sensitive to the target format, which is expected because GPQA differs from GSM8K and MATH-500 in both domain and answer style. Even in this harder setting, the transferred operator remains competitive with zero-shot inference for most model and task combinations, and it improves over the in-domain operator for one Llama model. These results suggest that TO extracts a partially task-general reasoning signal, while also preserving the ability to reflect task-specific structure when the source and target tasks are more closely aligned.

\begin{table}[h]
\centering
\caption{Cross-task transfer of task operators extracted from GSM8K to MATH-500 and GPQA-Diamond. The results compare transferred operators with zero-shot inference, in-domain task operators, and standard in-context learning. Underlined entries indicate methods that improve over the zero-shot baseline.}
\label{tab:transfer}
\small
\begin{tabular}{llcc}
\toprule
Model & Method & MATH-500 & GPQA-Diamond \\
\midrule
\multirow{4.5}{*}{\makecell{Qwen3\\(4B)}}
& \cellcolor{gray!15} ICL              & \cellcolor{gray!15} 55.80 & \cellcolor{gray!15} 37.88 \\
\cmidrule(lr){2-4}
& ZSL                                  & 23.60 & 17.68 \\
& TO (in-domain)                       & \underline{51.20} & \underline{33.33} \\
& TO (transfer)                        & \underline{51.20} & \underline{21.72} \\
\midrule
\multirow{4.5}{*}{\makecell{Qwen3\\(8B)}}
& \cellcolor{gray!15} ICL              & \cellcolor{gray!15} 61.20 & \cellcolor{gray!15} 34.85 \\
\cmidrule(lr){2-4}
& ZSL                                  & 19.00 & 21.21 \\
& TO (in-domain)                       & \underline{50.60} & \underline{33.33} \\
& TO (transfer)                        & \underline{41.80} & 20.71 \\
\midrule
\multirow{4.5}{*}{\makecell{Llama3.2\\(3B)}}
& \cellcolor{gray!15} ICL              & \cellcolor{gray!15} 27.20 & \cellcolor{gray!15} 24.24 \\
\cmidrule(lr){2-4}
& ZSL                                  & 24.80 & 26.26 \\
& TO (in-domain)                       & \underline{25.20} & 22.73 \\
& TO (transfer)                        & 24.20 & \underline{28.79} \\
\midrule
\multirow{4.5}{*}{\makecell{Llama3.1\\(8B)}}
& \cellcolor{gray!15} ICL              & \cellcolor{gray!15} 28.80 & \cellcolor{gray!15} 26.26 \\
\cmidrule(lr){2-4}
& ZSL                                  & 27.60 & 23.74 \\
& TO (in-domain)                       & 25.40 & 23.74 \\
& TO (transfer)                        & \underline{31.60} & 23.74 \\
\bottomrule
\end{tabular}
\end{table}

\section{Additional Results}
\label{sec:additional_results}

We provide additional results on ICV with a tuned steering strength, the inference efficiency of TO, and a normalized variant of TO for large numbers of demonstrations.

\subsection{ICV with Tuned Steering Strength}
\label{sec:additional_baselines}

In the main experiments, ICV follows its official implementation, which applies steering at all layers with $\lambda{=}0.1$.
To examine whether a task-specific steering strength improves ICV, we tune a single shared $\lambda \in \{0, 0.05, 0.1, 0.2, 0.4, 0.6, 0.8, 1.0\}$ for each task by minimizing the validation NLL, the same criterion used to select hyperparameters for the other baselines.
As shown in Table~\ref{tab:icv_tuned}, the tuned ICV performs similarly to the default setting and remains close to ZSL.

\begin{table}[h]
\centering
\caption{Accuracy (\%) of ICV with the default steering strength ($\lambda{=}0.1$) and with $\lambda$ tuned on the validation set.}
\label{tab:icv_tuned}
\small
\begin{tabular}{llcccc}
\toprule
Model & Task & ZSL & ICV ($\lambda{=}0.1$) & ICV (tuned $\lambda$) & TO (Ours) \\
\midrule
\multirow{2}{*}{Qwen3-4B} & Translation & 0.0 & 0.0 & 0.0 & 67.2 \\
 & Linguistic & 0.0 & 0.0 & 0.0 & 71.1 \\
\midrule
\multirow{2}{*}{Qwen3-8B} & Translation & 0.0 & 0.0 & 0.0 & 67.2 \\
 & Linguistic & 0.0 & 1.6 & 0.0 & 75.0 \\
\midrule
\multirow{2}{*}{Llama3.2-3B} & Translation & 0.0 & 0.0 & 0.8 & 68.0 \\
 & Linguistic & 3.9 & 4.7 & 4.7 & 75.8 \\
\midrule
\multirow{2}{*}{Llama3.1-8B} & Translation & 1.6 & 1.6 & 0.8 & 71.9 \\
 & Linguistic & 10.9 & 10.2 & 7.8 & 81.2 \\
\bottomrule
\end{tabular}
\end{table}

\subsection{Efficiency Analysis}
\label{sec:efficiency}

We quantify the per-query inference cost on 64 GSM8K questions with Llama3.1-8B, whose demonstrations are long chain-of-thought exemplars, assuming 128 generated tokens per query.
As shown in Table~\ref{tab:efficiency}, TO matches the cost of zero-shot inference, since it only rescales and shifts the attention outputs.
Compared with 8-shot ICL, it reduces the prompt length by more than $20\times$ and the key--value (KV) cache by more than $8\times$; caching the demonstrations avoids recomputing them for each query but still requires storing and attending to them.
The one-time extraction cost with $m{=}32$ validation prompts and $K{=}8$ demonstrations is about $32 \times 29{,}350$ GFLOPs, which is recovered after about 36 test queries relative to 8-shot ICL without KV caching.

\begin{table}[h]
\centering
\caption{Per-query inference cost on GSM8K with Llama3.1-8B, reported as mean $\pm$ standard deviation over 64 questions.}
\label{tab:efficiency}
\small
\begin{tabular}{lccc}
\toprule
Setting & Prompt tokens & KV cache (MB) & GFLOPs \\
\midrule
ZSL & 69$\pm$20 & 25$\pm$3 & 3180$\pm$320 \\
8-shot ICL & 1602$\pm$20 & 216$\pm$3 & 29350$\pm$330 \\
8-shot ICL + KV caching & 1602$\pm$20 & 216$\pm$3 & 3340$\pm$320 \\
TO (Ours) & 69$\pm$20 & 25$\pm$3 & 3180$\pm$320 \\
\bottomrule
\end{tabular}
\end{table}

\subsection{Normalized Task Operator}
\label{sec:nto}

As discussed in Section~\ref{sec:scaling_analysis}, the scaling factor $w = Z'/(Z' + Z_{\mathcal{C}})$, where $Z_{\mathcal{C}} = \sum_{i \in \mathcal{C}} \exp(s_i)$ is the softmax mass on context positions, approaches zero for large $K$ because $Z_{\mathcal{C}}$ grows with the number of demonstrations.
As an alternative to batch averaging, we consider a normalized task operator (NTO) that rescales the context mass to a reference number of demonstrations $K_0$:
\begin{equation*}
\tilde{w} = \frac{Z'}{Z' + (K_0/K)\, Z_{\mathcal{C}}}, \qquad \tilde{\mathbf{b}} = (1 - \tilde{w})\, \frac{\mathbf{b}}{1 - w},
\end{equation*}
which keeps the scaling factor comparable to that of a $K_0$-shot prompt while preserving the direction of the context contribution.
Table~\ref{tab:nto} compares NTO ($K_0{=}8$) with TO (batch) on Llama3.1-8B.
NTO further improves over TO (batch) on Deduplicate for all $K > 8$, whereas on Reverse it improves at $K{=}16$ but degrades for larger $K$.
As NTO does not consistently outperform the simpler batch averaging, we use TO (batch) in the main experiments.

\begin{table}[h]
\centering
\caption{Test accuracy (\%) on Llama3.1-8B as the number of demonstrations $K$ increases.}
\label{tab:nto}
\small
\begin{tabular}{llcccc}
\toprule
Task & Method & $K{=}8$ & $K{=}16$ & $K{=}32$ & $K{=}64$ \\
\midrule
\multirow{2}{*}{Deduplicate} & TO (batch) & 28.1 & 35.9 & 39.8 & 46.1 \\
 & NTO ($K_0{=}8$) & 28.1 & 38.3 & 42.2 & 51.6 \\
\midrule
\multirow{2}{*}{Reverse} & TO (batch) & 51.6 & 53.1 & 54.7 & 54.7 \\
 & NTO ($K_0{=}8$) & 51.6 & 61.7 & 53.1 & 43.0 \\
\bottomrule
\end{tabular}
\end{table}